\documentclass{article} % For LaTeX2e
\usepackage{iclr2027_conference,times}

\usepackage{amsmath,amssymb,amsthm,bm,bbm}
\usepackage{mathtools}
\usepackage{booktabs}
\usepackage{graphicx}
\usepackage{placeins}
\usepackage{wrapfig}
\usepackage{xcolor}
\usepackage{colortbl}
\newcommand{\faintrule}{\arrayrulecolor{black!25}\midrule\arrayrulecolor{black}}
\newcommand{\se}[1]{\textcolor{black!45}{\scriptsize$\,\pm#1$}}
\usepackage{tikz}
\usetikzlibrary{arrows.meta,calc}
\usepackage{algpseudocode}
\usepackage[colorlinks,linkcolor=red,citecolor=purple,urlcolor=blue]{hyperref}

\newcommand{\appref}[1]{\hyperref[#1]{Appendix~\ref*{#1}}}
\usepackage{url}
\usepackage{tcolorbox}
\tcbuselibrary{skins}
\usepackage{microtype}

\newtheorem{proposition}{Proposition}

\tcolorboxenvironment{proposition}{enhanced,colback=black!5,frame hidden,left=6pt,right=6pt,top=4pt,bottom=4pt}
\newcommand{\R}{\mathbb{R}}
\newcommand{\E}{\mathbb{E}}
\newcommand{\Cov}{\operatorname{Cov}}
\newcommand{\Var}{\operatorname{Var}}

\newcommand{\snr}{\operatorname{SNR}}
\newcommand{\x}{\mathbf{x}}
\newcommand{\w}{\mathbf{w}}
\newcommand{\z}{\mathbf{z}}
\newcommand{\va}{\mathbf{a}}
\newcommand{\vu}{\mathbf{u}}
\newcommand{\vphi}{\bm{\phi}}
\newcommand{\veta}{\bm{\eta}}
\newcommand{\vtheta}{\bm{\theta}}

\newcommand{\mI}{\bm{I}}

\newcommand{\eps}{\bm{\varepsilon}}

\newcommand{\kl}{\operatorname{KL}}
\newtcolorbox{examplebox}[1]{colback=blue!3,colframe=blue!50!black,title=#1,fonttitle=\bfseries,boxrule=0.5pt,left=4pt,right=4pt,top=2pt,bottom=2pt}
\newtcolorbox{takeaway}{colback=orange!5,colframe=orange!70!black,boxrule=0.5pt,left=4pt,right=4pt,top=2pt,bottom=2pt}
\newtcolorbox{keyidea}[1]{
    enhanced,
    colback=white,
    colframe=black!35,
    boxrule=0.6pt,
    colbacktitle=white,
    coltitle=black,
    fonttitle=\bfseries,
    title=#1,
    attach boxed title to top left={xshift=0.6cm,yshift=-3mm},
    boxed title style={frame hidden,boxrule=0pt},
    left=6pt,
    right=6pt,
    top=7pt,
    bottom=5pt
}

\title{Reward Tilting via Semantic Noising}

\hypersetup{colorlinks=true,linkcolor=red!70!black,citecolor=blue!50!green,urlcolor=blue}

\author{
Rafał Karczewski\thanks{Correspondence to \texttt{rafal.karczewski@aalto.fi} and \texttt{vgarg@csail.mit.edu}.}\textsuperscript{\phantom{*},}\thanks{Now at Microsoft Research.}\\
Aalto University
\And
Peter Blohm\\
Technische Universität Wien
\And
Markus Heinonen\thanks{Now at basis.ai.}\\
Aalto University
\And
Vikas Garg\\
Aalto University\\YaiYai Ltd
}

\iclrpreprint
\begin{document}

\maketitle

\begin{abstract}
Steering a generative model toward a desired property typically requires separate fine-tuning for each reward or additional computation during sampling. We propose {\em semantic noising}, which conditions generation on a noisy semantic representation of the data. The resulting posteriors correspond to exponential tilts by linear-quadratic rewards in semantic space, allowing a single model to support objectives and strengths chosen after training. A conditional generator can be trained to approximate these posteriors directly, while models that generate semantic representations by diffusion provide analytic tilted denoisers for the same family of rewards. We illustrate semantic noising in three settings. An autoregressive molecular generator trained with semantic noising controls properties individually and jointly, with a better reward--KL trade-off than property-conditioned models. On ImageNet, a joint diffusion model of images and their CLIP embeddings achieves FID comparable to REG, and tilting it toward a higher aesthetic score matches the best of 16 samples at the cost of a single sample. Finally, pretrained factorized models such as Kandinsky 2.2 support the same tilts without retraining.
\end{abstract}

\section{Introduction}
In many applications, we want a generative model to favor particular properties, such as greater aesthetic quality in images or a desired molecular attribute. The objective and its desired strength are often chosen only after the model has been trained. Existing approaches commonly fine-tune the model separately for each reward \citep{pmlr-v202-lu23d,liu2025flowgrpo,domingo-enrich2025adjoint,bergmeister2026reinforceadjointmatchingscaling,potaptchik2026tilt} or add computation during sampling \citep{skreta2025feynmankac,singhal2025a}, making it costly to accommodate different objectives. We aim to support a useful family of rewards with a single model, without repeating this work for each reward.

For a given reward function $r(\x)$, we seek a distribution that favors higher-reward examples while staying close to the base distribution $q(\x)$. The distribution that maximizes the expected reward penalized by the KL divergence from $q$ is the exponential tilt \citep{donsker1983asymptotic},
\[
    q^r(\x)\propto q(\x)\exp r(\x).
\]
Efficient sampling from $q^r$ is generally difficult for arbitrary $r$.
\begin{figure}[t]
    \centering
    \includegraphics[width=\linewidth]{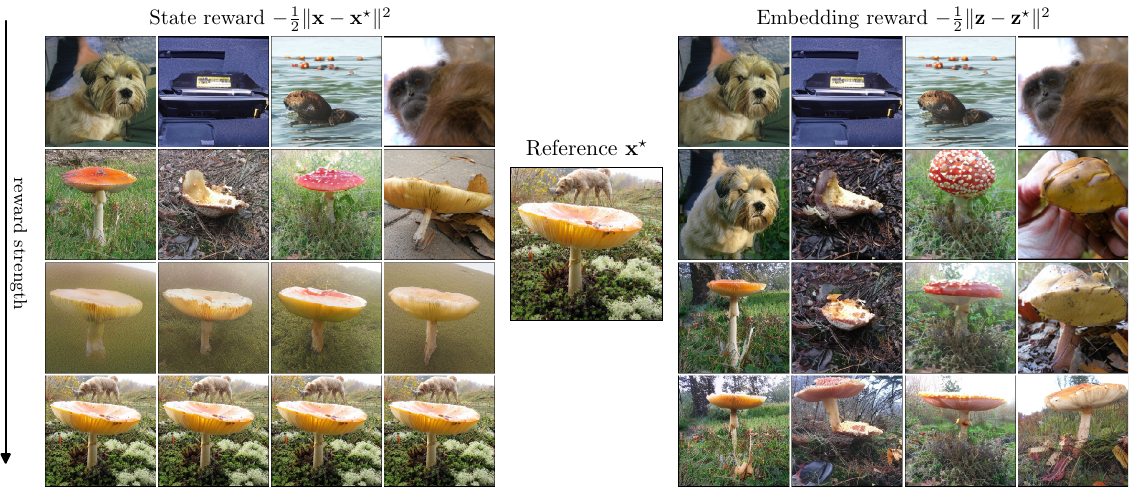}
    \caption{\textbf{Tilting toward a reference in semantic space.} Samples from our joint diffusion model (\autoref{sec:joint-exp}) tilted toward a reference image $\x^\star$ (center) using the pixel-space reward $-\tfrac{\lambda}{2}\|\x-\x^\star\|^2$ (left) or the CLIP-space reward $-\tfrac{\lambda}{2}\|\z-\z^\star\|^2$ with $\z^\star=\vphi(\x^\star)$ (right), with $\lambda$ increasing from top to bottom. The pixel-space reward pulls samples toward the reference image itself, whereas the CLIP-space reward favors diverse images with similar semantic content.}
    \label{fig:state-vs-embedding}
\end{figure}
However, many useful objectives take a simple form in a suitable representation of the data, which we call the \emph{semantic space}. For example, a reward that favors proximity in this space prefers images with similar content even when their pixels differ (\autoref{fig:state-vs-embedding}), while a linear projection can predict properties such as aesthetic quality or molecular attributes. We show how to efficiently tilt by such rewards via \emph{semantic noising}, which conditions generation on a noisy semantic representation. By Bayes' rule, conditioning on the noisy representation reweights $q$ by a Gaussian likelihood, which is the exponential of a reward that is linear-quadratic in semantic space. The choice of representation therefore defines the reward family, while the noisy representation and its noise level specify a particular reward within it.

We describe two ways to sample from these tilted distributions. First, a conditional generative model can learn the posteriors directly from examples paired with their noisy representations. Second, in diffusion models that generate semantic representations, the tilted denoiser is obtained by evaluating the original denoiser at shifted inputs, without additional denoiser evaluations \citep{liu2024flow,holderrieth2026glass,moitra2026quadratic,mcdonald2026technical}. Models that jointly diffuse data and semantic representations \citep{kouzelis2025boosting,wu2025representation} support these rewards once the data and semantic noise levels are sampled independently during training. Factorized models such as unCLIP and Kandinsky 2.2 \citep{ramesh2022hierarchical,razzhigaev2023kandinsky}, where a diffusion prior generates the semantic representation and a separate decoder generates the data, support them without retraining.

We evaluate both approaches in molecular and image generation. We train MolGPT \citep{molgpt}, an autoregressive molecular generator, to condition on noisy CLAMP embeddings \citep{pmlr-v202-seidl23a}, and control properties individually and jointly through linear predictors on these embeddings, with a better reward--KL trade-off than separately trained property-conditioned models. On ImageNet \citep{imagenet}, our joint diffusion model achieves FID comparable to REG \citep{wu2025representation}. With moderate tilts, it matches the aesthetic reward of the best of 16 samples and of diffusion posterior sampling \citep{chung2023diffusion} using an order of magnitude less GPU time, and scores higher than both under two other aesthetic predictors. We also compare with Feynman--Kac steering \citep{singhal2025a}, which generates more realistic images for stronger tilts when it uses the semantic representation predicted by our model, but also costs an order of magnitude more than our method. Finally, we show that a pretrained Kandinsky 2.2 model supports these tilts without retraining or additional denoiser evaluations.

\section{Reward tilting via semantic noising}\label{sec:semantic-noising}
Let $q(\x)$ be the data distribution. The {\em exponential tilt} by a reward $r(\x)$ is the unique solution of KL-regularized reward maximization,
\begin{equation}\label{eq:tilt-objective}
    q^r \coloneqq \arg\max_p \left\{\E_p[r(\x)]-\kl[p\|q]\right\} \quad \Longrightarrow \quad  q^r(\x)\propto q(\x)\exp r(\x).
\end{equation}
We consider rewards defined on a semantic representation $\z=\vphi(\x) \in \R^k$, where $\vphi$ is a fixed feature map. The features may come from a pretrained encoder, such as CLIP \citep{pmlr-v139-radford21a} for images, or be hand-designed. Our construction relates a family of rewards in semantic space to conditioning on Gaussian-corrupted representations.

\subsection{Posteriors induced by semantic noising}\label{sec:noisy-posteriors}
We define \emph{semantic noising} by adding Gaussian noise to the semantic representation. For a noise precision $\tau_\z>0$, we use the noisy representation $\vphi(\x)+\eps/\sqrt{\tau_\z}$ scaled by $\tau_\z$,
\begin{equation}\label{eq:sem-noise-obs}
    \tilde{\z}=\tau_\z\vphi(\x)+\sqrt{\tau_\z}\,\eps,
    \qquad \eps\sim\mathcal{N}(\bm{0},\mI).
\end{equation}
Expanding the Gaussian likelihood and applying Bayes' rule identifies the posterior as an exponential tilt with a linear-quadratic semantic reward. For any $\va \in \R^k$ and $\gamma>0$, we have
\begin{keyidea}{Semantic noising as reward tilting}
\begin{equation}\label{eq:sem-noise-posterior}
    q(\x\mid\tilde{\z}=\va,\tau_\z=\gamma)
    \propto q(\x)\exp\!\Bigl(
        \va\cdot\vphi(\x)
        -\tfrac{\gamma}{2}\|\vphi(\x)\|^2
    \Bigr) = q(\x)\exp\! \left(r_{\va,\gamma}(\x)\right).
\end{equation}
\end{keyidea}
The feature map $\vphi$ determines which rewards can be expressed, and the conditioning parameters $(\tilde{\z},\tau_\z)$ select a member of this family, with $(\tilde{\z},\tau_\z)=(\bm{0},0)$ recovering $q$. We use two kinds of rewards from this family throughout the paper. The first favors proximity to a reference $\z^\star$ in semantic space. For a strength $\lambda$, the reward $-\tfrac{\lambda}{2}\|\vphi(\x)-\z^\star\|^2$ equals $r_{\va,\gamma}$ up to an additive constant for $\va=\lambda\z^\star$ and $\gamma=\lambda$. The second is a linear predictor $\va\cdot\vphi(\x)$ of a property such as aesthetic quality or a molecular attribute. A weighted sum of linear predictors is again a linear predictor, so several properties can be controlled simultaneously without fitting new predictors.

If $\|\vphi(\x)\|$ is constant, as for the normalized embeddings in \autoref{sec:molgpt-exp} and \autoref{sec:joint-exp}, the quadratic term is constant across examples and does not affect the tilted distribution. The family then reduces to rewards that are linear in $\vphi(\x)$, and the posterior no longer depends on $\tau_\z$ once $\tilde{\z}$ is fixed, i.e., $q(\x\mid \tilde \z,\tau_\z)=q(\x\mid \tilde \z)$.

A reward $r$ that is not linear-quadratic in $\vphi$ can still be approximated by some $\hat r=r_{\va,\gamma}$, for example a linear predictor of $r$ fitted by regression, and the quality of this approximation determines how close the tilt by $\hat r$ is to the tilt by $r$.
\begin{proposition}[Tilting with an approximate reward]\label{prop:approx-reward}
If $r(\x)$ and $\hat r(\x)$ are jointly Gaussian for $\x\sim q$, then $\kl(q^{\hat r}\,\|\,q^{r})=\kl(q^{r}\,\|\,q^{\hat r})=\tfrac{1}{2}\Var_q[r-\hat r]$ and
\begin{equation}\label{eq:approx-reward-r2}
    \kl\bigl(q^{\hat r}\,\|\,q^{r}\bigr)=(1-R^2)\,\kl\bigl(q\,\|\,q^{r}\bigr),
\end{equation}
where $R^2=1-\Var_q[r-\hat r]/\Var_q[r]$ is the coefficient of determination of $\hat r$.
\end{proposition}
We prove \autoref{prop:approx-reward} and generalize it beyond the Gaussian case in \appref{app:approx-reward}.

\subsection{Direct conditional generation}\label{sec:direct-cond-new}
\autoref{eq:sem-noise-posterior} suggests a direct way to learn the tilted distributions. A generative model trained with its usual objective on examples $\x\sim q$ approximates $q(\x)$. If each training example is paired with a noisy representation $\tilde{\z}$ from \autoref{eq:sem-noise-obs} at a random precision $\tau_\z$, and the pair $(\tilde{\z},\tau_\z)$ is given to the model as a conditioning input, the same objective yields a conditional model
\begin{equation}\label{eq:direct-cond-train}
    p_{\vtheta}(\x\mid\tilde{\z},\tau_\z)
    \approx q(\x\mid\tilde{\z},\tau_\z).
\end{equation}
Training is otherwise unchanged, and the reward enters only through the conditioning input at inference time. To sample from the tilt by $r_{\va,\gamma}$ with $\gamma>0$, we set $\tilde{\z}=\va$ and $\tau_\z=\gamma$, so that $p_{\vtheta}(\x\mid\tilde{\z}=\va,\tau_\z=\gamma) \approx q(\x\mid\tilde{\z}=\va,\tau_\z=\gamma) \propto q(\x)\exp(r_{\va,\gamma}(\x))$. We illustrate training and sampling for an autoregressive model in \appref{app:algorithms}.

Linear rewards correspond to $\gamma=0$, which a trained model cannot reliably represent, because small $\tau_\z$ gives $\tilde\z\approx\bm{0}$ in \autoref{eq:sem-noise-obs}, and the model can learn to ignore $\tilde\z$ in this regime. However, if $\|\vphi(\x)\|$ is constant, the posterior does not depend on $\tau_\z$ (\autoref{sec:noisy-posteriors}), so the model can be conditioned on $\tilde\z$ alone. A bounded representation can be lifted to a sphere by adding one coordinate (\appref{app:exact-rewards}).

We demonstrate direct conditional generation by training an autoregressive molecular generator in \autoref{sec:molgpt-exp}. Existing models trained to condition on noisy semantic representations also support these tilts without retraining, for example Stable unCLIP \citep{Rombach_2022_CVPR,rombach2023stableunclip}, which conditions an image diffusion model on noisy CLIP image embeddings (\appref{app:unclip}).

\section{Analytic tilting with semantic diffusion}\label{sec:analytic-diffusion}
Diffusion models are a natural fit for semantic noising, since each sampling step evaluates a denoiser at a noisy version of the generated variable. When a diffusion model generates the semantic representation, either jointly with the data or as a separate prior, the tilts of \autoref{sec:semantic-noising} can be obtained from its denoiser without training a separate conditional model.

\subsection{Primer on diffusion models and natural coordinates}\label{sec:diffusion-natcoord-new}
A Gaussian diffusion model for a data distribution $q(\x)$ defines noisy variables $\x_t=\alpha_t\x+\sigma_t\eps$ for $t\in[0,T]$, where the signal-to-noise ratio $\snr(t)=\alpha_t^2/\sigma_t^2$ decreases with $t$ \citep{song2020score,ho2020denoising,lipman2023flow,albergo2023building,liu2023flow}. Samples are generated by integrating a reverse SDE or the deterministic probability-flow ODE (PF-ODE) from $\x_T\sim\mathcal{N}(\bm{0},\sigma_T^2\mI)$ to $t=0$, using the score $\nabla\log p_t$ of the noisy marginals approximated by a model \citep{song2020score,JMLR:v6:hyvarinen05a,6795935}. By Tweedie's formula \citep{efron2011tweedie}, the score satisfies $\sigma_t^2 \nabla \log p_t(\x_t) = \alpha_t \E[\x \mid \x_t, t] - \x_t$. The generative process therefore depends on $q$ only through the denoiser $D(\x_t,t)=\E[\x\mid\x_t,t]$.

\citet{karczewski2025spacetime} observed that the denoising posteriors form an exponential family with natural coordinates
\begin{equation}\label{eq:natural-coordinates-new}
    \tilde \x = \frac{\alpha_t \x_t}{\sigma_t^2}, \quad \tau = \frac{\alpha_t^2}{\sigma_t^2}.
\end{equation}
In these coordinates, we have $\tilde{\x}=\tau\x+\sqrt{\tau}\,\eps$, which has the same form as semantic noising in \autoref{eq:sem-noise-obs}. Unlike the standard coordinates $(\x_t,t)$, the natural coordinates do not depend on the choice of noise schedule, and they coincide with the coordinates of the associated stochastic localization process \citep{montanari2023sampling}. We write $D(\tilde{\x},\tau)\coloneqq\E[\x\mid\tilde{\x},\tau]$ for the denoiser in these coordinates.

In practice, natural coordinates are numerically unstable near $t=0$, so we parametrize models in standard coordinates. These are recovered from $(\tilde{\x},\tau)$ by $t=\snr^{-1}(\tau)$ and $\x_t=\alpha_t\tilde{\x}/\tau$. We provide further background in \appref{app:exp-families} and \appref{app:exp-diffusion}.

\subsection{Linear-quadratic tilting}\label{sec:lq-tilting-new}
For a tilted data distribution $q^r$, the corresponding denoiser is given by
\begin{equation*}
    D^{r}(\tilde \x,\tau) = \frac{\E\big[\x \exp r(\x) \mid \tilde \x,\tau\big]}{\E \big[\exp r(\x) \mid \tilde \x,\tau\big]},
\end{equation*}
which is generally intractable. For a linear-quadratic reward, however, multiplying the denoising posterior by $\exp r(\x)$ shifts its natural parameters, so the tilted denoiser satisfies \citep{mcdonald2026technical}
\begin{equation}\label{eq:general-denoiser-new}
    r(\x) = \va\cdot \x - \tfrac{\gamma}{2}\|\x\|^2
    \quad \Longrightarrow \quad
    D^{r}(\tilde{\x}, \tau) = D(\tilde{\x} + \va, \tau + \gamma),
\end{equation}
for $\gamma\geq0$. This includes the purely linear case $\gamma=0$ \citep{moitra2026quadratic} and the quadratic proximity reward $-\tfrac{\lambda}{2}\|\x-\x^\star\|^2$, which corresponds to $\va=\lambda\x^\star$ and $\gamma=\lambda$ \citep{liu2024flow}. Substituting this denoiser into the reverse SDE or PF-ODE gives a sampling process for the tilted distribution, which uses the same number of denoiser evaluations as standard sampling, only at shifted inputs. We provide a proof in natural coordinates and discuss the connection to posterior sampling in \appref{app:linear-tilt}.

Although \autoref{eq:general-denoiser-new} holds for $\gamma=0$, this case is problematic for a learned denoiser. The shifted input then corresponds to $\x_t+(\sigma_t^2/\alpha_t)\va$ in standard coordinates, and relative to $\x_t$ the shift grows as $\tau^{-1/2}$ when $\tau\to0$, so near $t=T$ the denoiser is evaluated far from the inputs it was trained on, for any noise schedule. When the diffused variable has constant norm, as for the normalized CLIP embeddings in \autoref{sec:joint-exp}, every $\gamma>0$ gives the tilt by the same linear reward (\autoref{sec:noisy-posteriors}), so we apply linear rewards with a small $\gamma>0$.

\subsection{Joint diffusion of data and semantic representations}\label{sec:joint-diffusion-new}
Several models jointly diffuse data and their semantic representations \citep{kouzelis2025boosting,kouzelis_coredi2026,wu2025representation}. We show that such models also support the tilts of \autoref{sec:semantic-noising} if the data and semantic noise levels are sampled independently during training.

Let $q(\x,\z)$ be the joint distribution with $\z=\vphi(\x)$. We noise the two coordinates independently,
\begin{equation*}
    \x_{t_\x} = \alpha_{t_\x} \x + \sigma_{t_\x} \eps_\x, \quad \z_{t_\z} = \alpha_{t_\z} \z + \sigma_{t_\z} \eps_\z,
\end{equation*}
with independent diffusion times $t_\x,t_\z$ and independent noise $\eps_\x,\eps_\z\sim\mathcal{N}(\bm{0},\mI)$. Each coordinate has its own natural parameters $(\tilde\x,\tau_\x)$ and $(\tilde\z,\tau_\z)$, giving a joint denoiser $(D_\x,D_\z)(\tilde\x,\tau_\x,\tilde\z,\tau_\z)$. For the semantic reward $r_{\va, \gamma}$ (\autoref{eq:sem-noise-posterior}), the tilted joint denoiser shifts only the semantic inputs (\appref{app:joint-quadratic-tilt}),
\begin{keyidea}{Joint semantic tilting}
\begin{equation}\label{eq:joint-tilted-denoiser-new}
    (D_\x^{r}, D_\z^{r})(\tilde \x,\tau_\x,\tilde \z,\tau_\z) = (D_\x, D_\z)\big(\tilde \x,\tau_\x, \tilde \z+\va, \tau_\z+\gamma \big).
\end{equation}
\end{keyidea}
Sampling with the shifted denoiser targets the joint distribution proportional to $q(\x,\z)\exp r(\z)$, whose data marginal is the tilted distribution of \autoref{sec:semantic-noising}.

In a model trained with a shared noise level, we always have $\tau_\x=\tau_\z$ during training, whereas \autoref{eq:joint-tilted-denoiser-new} shifts $\tau_\z$ while leaving $\tau_\x$ unchanged. Training must therefore cover different combinations of the two noise levels, which is why we sample them independently.

\subsection{Semantic diffusion priors}\label{sec:embedding-only-new}
Instead of jointly diffusing data and semantic representations, we can use a semantic diffusion prior and a separate decoder that generates the data given the semantic representation, i.e., $q(\x,\z)=q(\z)q(\x\mid\z)$. Because the reward depends only on $\z$, the tilted distribution factorizes as
\begin{equation*}
    q^r(\x,\z) = q^r(\z)q(\x\mid\z),
\end{equation*}
where $q^r(\z)\propto q(\z)\exp r(\z)$. Sampling then draws $\z \sim q^r(\z)$ and generates data from the unchanged decoder $\x \sim q(\x\mid\z)$. If $q(\z)$ is modeled by diffusion, its tilted denoiser follows directly from \autoref{eq:general-denoiser-new}, whereas the decoder need not be a diffusion model. Models such as unCLIP \citep{ramesh2022hierarchical} and Kandinsky 2.2 \citep{razzhigaev2023kandinsky} have the structure of a semantic diffusion prior and a separate image decoder. Applying the shift only to the prior gives semantic tilting without retraining either component or increasing the number of denoiser evaluations.

\subsection{Feynman--Kac steering for strong tilts}\label{sec:semantic-fks}
\autoref{eq:joint-tilted-denoiser-new} is exact for the true denoiser, but it evaluates the learned denoiser at shifted inputs, which may differ from the inputs seen during training, especially for strong tilts. Feynman--Kac steering (FKS; \citealp{singhal2025a}) avoids this problem at the cost of additional computation during sampling. It keeps track of several particles and resamples them according to an estimate of their final reward, typically the reward of the denoised data, $r(D(\tilde\x,\tau))$, or its average over samples from the denoising posterior, which requires additional denoiser evaluations. For a reward that is linear in the semantic representation, $r(\z)=\va\cdot\z$, the expected final reward follows from the denoised semantic representation without additional denoiser evaluations,
\begin{equation}\label{eq:semantic-fks}
    \E\bigl[\va\cdot\z\mid\tilde\x,\tau_\x,\tilde\z,\tau_\z\bigr]=\va\cdot D_\z(\tilde\x,\tau_\x,\tilde\z,\tau_\z).
\end{equation}
We call FKS with this estimate \emph{semantic FKS}.\footnote{Semantic FKS does not require the noisy semantic representation as an input, so it could also be applied to models that predict the denoised semantic representation from $(\tilde\x,\tau_\x)$ alone.}

\section{Experiments}
\label{sec:exp}

\subsection{Post-hoc property control with MolGPT}\label{sec:molgpt-exp}
CLAMP \citep{pmlr-v202-seidl23a} maps each molecule to a 768-dimensional embedding, which we normalize to obtain $\vphi(\x)$. Because $\|\vphi(\x)\|=1$, we have $q(\x\mid\tilde{\z},\tau_\z)=q(\x\mid\tilde{\z})$ (\autoref{sec:noisy-posteriors}). We model $q(\x\mid\tilde{\z})$ with MolGPT \citep{molgpt}, an autoregressive SMILES generator initialized from the released unconditional GuacaMol checkpoint. During training, we sample $(\tilde{\z},\tau_\z)$ for each molecule according to \autoref{eq:sem-noise-obs}, condition the model on $\tilde{\z}$ through a learned prefix, and use the standard next-token cross-entropy loss (\appref{app:molgpt}).

Independently of generator training, we fit linear predictors of ring count, molecular weight, rotatable bonds, topological polar surface area (TPSA), logarithm of octanol-water partition coefficient (logP), and synthetic accessibility score (SAS) from normalized CLAMP embeddings. For a predictor $\widehat R(\x)=\mathbf{w}\cdot\vphi(\x)+b$, we tilt by the reward $c\widehat R(\x)/s_{\widehat R}$, where $s_{\widehat R}$ is the standard deviation of the predicted property and $c$ sets the strength, by conditioning on $\tilde{\z}=c\mathbf{w}/s_{\widehat R}$.

\begin{figure}[t]
    \centering
    \includegraphics[width=\linewidth]{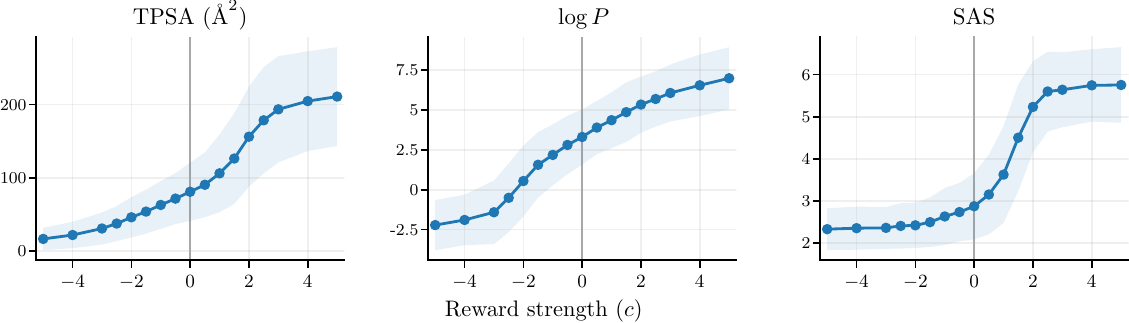}
    \caption{\textbf{A single generator controls multiple properties chosen after training.}
    The generator is trained to condition on noisy CLAMP embeddings, and property labels are used only to fit the post-hoc linear predictors that define the conditioning vectors. Shaded regions show one standard deviation among valid molecules. More properties in \autoref{fig:molgpt-conditioning-full}.}
    \label{fig:molgpt-conditioning}
\end{figure}

\autoref{fig:molgpt-conditioning} shows control of multiple molecular properties over $-5\leq c\leq5$, with results for more properties in \appref{app:molgpt}. The mean properties change monotonically with $c$.

\begin{figure}[t]
    \centering
    \includegraphics[width=\linewidth]{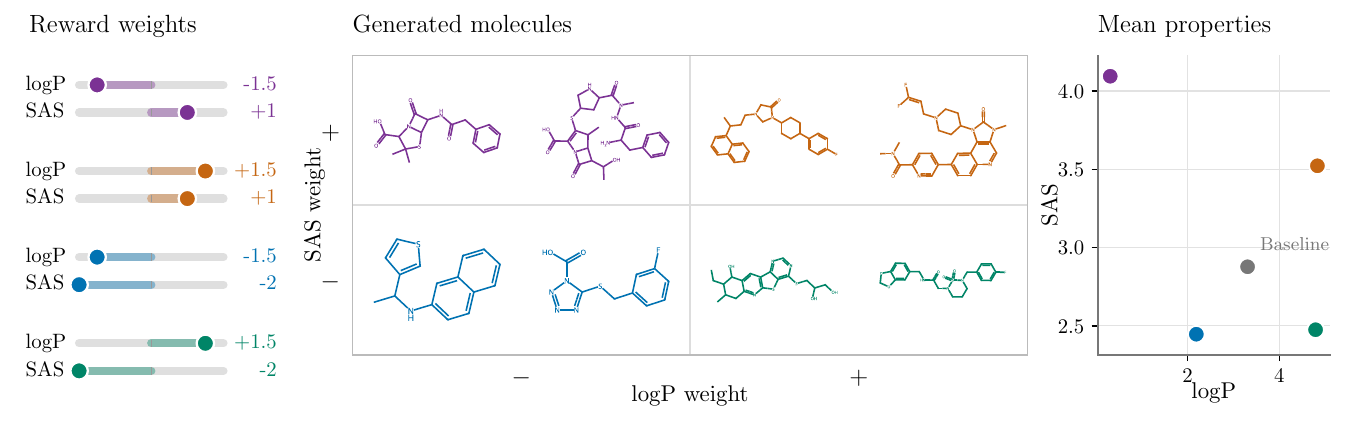}
    \caption{\textbf{Simultaneous control of multiple properties.}
    Left: requested control directions for SAS and logP. Middle: selected generated molecules for each setting. Right: mean properties in each setting compared to the untilted baseline.}
    \label{fig:molgpt-joint}
\end{figure}

\paragraph{Simultaneous control of multiple properties.}
Since a weighted sum of linear rewards is again a linear reward (\autoref{sec:noisy-posteriors}), we control logP and SAS jointly with the same model by conditioning on a weighted sum of their standardized predictor directions. For each tested pair of weights, both mean properties move in the requested directions relative to the untilted baseline (\autoref{fig:molgpt-joint}).

\begin{figure}[t]
    \centering
    \includegraphics[width=\linewidth]{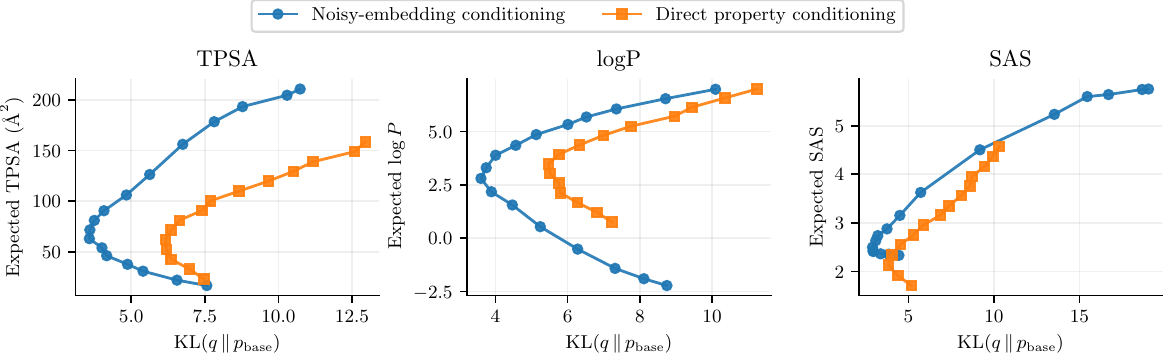}
    \caption{\textbf{Noisy-embedding conditioning gives a better reward--KL trade-off than direct property conditioning.}
    Direct property conditioning uses a separate MolGPT model for each property. Each panel compares the mean property value with the KL divergence from the same unconditional model. Curves further to the left are closer to the optimal trade-off attained by the exponential tilt.}
    \label{fig:molgpt-reward-kl}
\end{figure}

\paragraph{Comparison with property conditioning.}
We compare this model with three MolGPT models trained with the same budget to condition on TPSA, logP, or SAS, following the original MolGPT setup. By \autoref{eq:tilt-objective}, among all distributions with a given mean property, an exponential tilt has the smallest KL divergence from the base distribution. Plotting the mean property against this KL divergence therefore compares the models by how close they come to the optimal trade-off. Conditioning on a property value does not produce a tilt, and it reaches the same mean property values at higher KL divergence than noisy-embedding conditioning (\autoref{fig:molgpt-reward-kl}).

\paragraph{Sources of approximation error.}
The generated molecules differ from the tilt by the actual property for two reasons. First, the model only approximates the tilt by the predicted property. Since MolGPT gives exact likelihoods, we can estimate its KL divergence from this tilt, which is about half that of the untilted model (\appref{app:exp-tilt-accuracy}). Second, the CLAMP-based predictors approximate TPSA, logP, and SAS with $R^2$ between 0.75 and 0.80 (\autoref{tab:molgpt-readouts}). If the predicted and actual properties are jointly Gaussian, the KL divergence from the tilt by a predicted property to the tilt by the property itself is then 20 to 25\% of that of the untilted distribution (\autoref{prop:approx-reward}). Using the properties themselves as the semantic representation did not improve the reward--KL trade-off (\appref{app:exact-rewards}).

\subsection{Joint diffusion on ImageNet}\label{sec:joint-exp}
\begin{wrapfigure}[18]{r}{0.47\textwidth}
    \centering
    \vspace{-1em}
    \includegraphics[width=\linewidth]{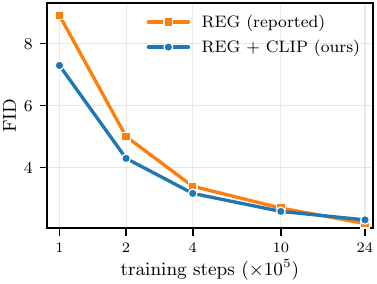}
    \caption{\textbf{FID over training.} Our model improves faster early in training, while REG is slightly better after roughly 2M steps.}
    \label{fig:reg_fid}
\end{wrapfigure}
We evaluate the joint diffusion construction on $256\times256$ ImageNet \citep{imagenet}. We closely follow REG \citep{wu2025representation}, a SiT-XL/2 model \citep{ma2024sit} that augments the image tokens with a DINOv2 \citep{oquab2024dinov} embedding token. We retain the DINOv2 token and add a normalized CLIP \citep{pmlr-v139-radford21a} embedding token. In the notation of \autoref{sec:joint-diffusion-new}, the data coordinate $\x$ consists of the image latent and its DINOv2 token, which share a noise level, and the semantic coordinate $\z$ is the normalized CLIP embedding, whose noise level is sampled independently. The additional token adds less than 1\% to the per-update training cost of REG with cached CLIP targets and to the cost of each denoiser evaluation (\appref{app:scaling}). We chose this design in an ablation study at a smaller scale (\appref{app:ablation}).

Following the sampling protocol of REG, we use 250 Euler--Maruyama steps without classifier-free guidance to generate 50,000 images and track FID \citep{NIPS2017_8a1d6947} over training (\autoref{fig:reg_fid}). Our model has slightly lower FID through 1M training steps, but then begins to saturate, and REG becomes better after roughly 2M steps. At 2.4M steps, FID is 2.32 for our model compared to the reported value of 2.2 for REG.

\begin{figure}[tbh]
    \centering
    \includegraphics[width=\linewidth]{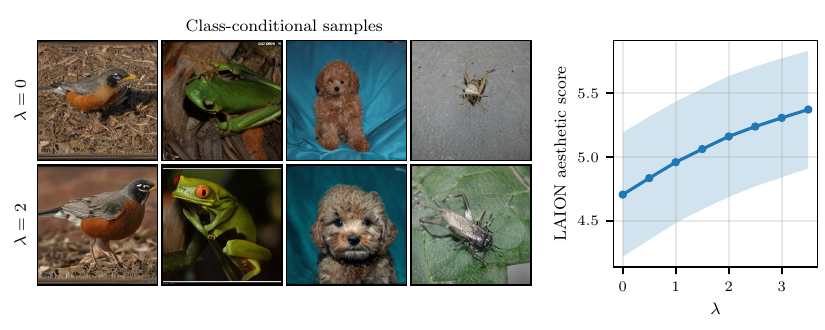}
    \caption{\textbf{Aesthetic tilting monotonically improves aesthetic score.} Selected class-conditional ImageNet samples at $\lambda=0$ (top) and $\lambda=2$ (bottom). Right: LAION aesthetic score against tilt strength (mean $\pm$ std over 1,536 samples from 24 classes).}
    \label{fig:aesthetic-results}
\end{figure}

\paragraph{Aesthetic-score tilting.} The LAION aesthetic score predictor \citep{schuhmann2022laionb}, which estimates human-perceived aesthetic quality, is a linear projection $S(\z)$ of the normalized CLIP embedding. Let $\sigma_S$ be the standard deviation of $S$ over untilted samples and $\va$ the weight vector of $S/\sigma_S$, so that $\va\cdot\z=S(\z)/\sigma_S$ up to a constant. Since $\|\z\|=1$, we apply this linear reward with a small quadratic term (\autoref{sec:lq-tilting-new}), using $r(\z) = \lambda(\va \cdot \z - \tfrac{\gamma\|\va\|}{2}\|\z\|^2)$, which shifts the embedding coordinates to $(\tilde \z + \lambda \va,\; \tau_\z + \lambda\gamma\|\va\|)$. With $\gamma=0.01$, the aesthetic score increases monotonically with $\lambda$ (\autoref{fig:aesthetic-results}). In theory, $\gamma$ does not change the target distribution, but it affects the output of the learned denoiser, so we treat it as a hyperparameter (\appref{app:gamma-ablation}).

\paragraph{Comparison with inference-time steering.} We compare aesthetic tilting with Best-of-$N$ sampling, FKS and diffusion posterior sampling (DPS; \citealp{chung2023diffusion}). All methods use the same joint model, and FKS, DPS and our method target $p(\x)\exp(\lambda S/\sigma_S)$. The baselines compute $S$ from decoded images, and our method from the generated CLIP embedding (\appref{app:realism}). At intermediate steps, FKS scores the decoded denoised image or, in semantic FKS, the denoised CLIP embedding (\autoref{eq:semantic-fks}). DPS adds the gradient of the reward of the decoded denoised image to the score at every step, which requires backpropagating through CLIP, the VAE decoder and the denoiser. To check that the gains are not specific to the LAION predictor, we also evaluate MUSIQ \citep{ke2021musiq} and NIMA \citep{talebi2018nima}, two aesthetic predictors that do not use CLIP. We measure the cost of steering by realism rather than by the KL divergence from the base model, since realism is easier to interpret and can be estimated from samples alone for every method. We use precision \citep{kynkaanniemi2019improved}, the fraction of generated images whose DINOv2 features \citep{oquab2024dinov} lie close to those of a real image of the same class, and report diversity in \appref{app:steering}.

\begin{figure}[tbh]
    \centering
    \includegraphics[width=\linewidth]{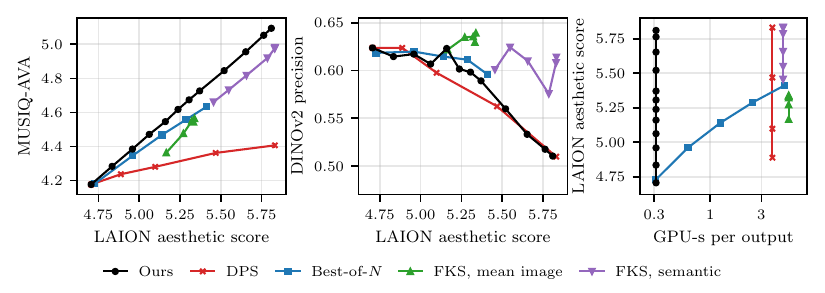}
    \caption{\textbf{Aesthetic steering on ImageNet.} MUSIQ score (left) and precision (middle) against the LAION score, and the LAION score against GPU time per output (right). Means over 1,536 outputs, with standard errors of about 0.013 for MUSIQ and 0.012 for precision.}
    \label{fig:steering}
\end{figure}

\autoref{fig:steering} shows that our method at $\lambda=3.5$ matches the LAION score and precision of Best-of-16 at 16 times less GPU time. Up to this strength, it follows the exact tilt of the learned model, which we estimate by reweighting untilted samples (\appref{app:realism}). At the same LAION score, our method is higher in MUSIQ and NIMA than every baseline (\appref{app:steering}). Stronger tilts raise all three aesthetic scores further but lower the precision. DPS costs 12 times as much as our method and has a similar precision at the same LAION score, whereas FKS costs 15 to 17 times as much and keeps a higher precision, especially semantic FKS. Universal guidance \citep{bansal2024universal} adds self-recurrence to DPS, which raises its precision. Self-recurrence raises the precision of our method by a similar amount, at about 13 times less cost than universal guidance (\appref{app:recurrence}).

\begin{figure}[tbh]
    \centering
    \includegraphics[width=\linewidth]{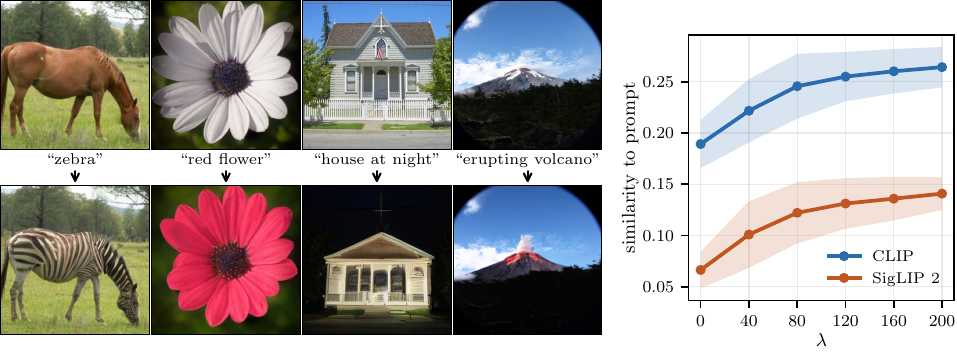}
    \caption{\textbf{Prompt-based image editing via semantic tilting.} Source images (top) edited by tilting toward the CLIP text embedding of a prompt (bottom). Right: CLIP and SigLIP 2 similarity between edited images and prompt increases monotonically with $\lambda$ (mean $\pm$ std over 42 source images).}
    \label{fig:prompt-editing}
\end{figure}

\paragraph{Prompt-based editing.} Given a text prompt, let $\z^\star(\mathrm{prompt})$ denote its CLIP text embedding and define the reward $-\tfrac{\lambda}{2}\|\z - \z^\star(\mathrm{prompt})\|^2$, corresponding to $\va=\lambda\z^\star(\mathrm{prompt})$ and $\gamma=\lambda$. For a given image $\x$, we compute its CLIP embedding $\z=\vphi(\x)$ and encode $(\x,\z)$ by running the PF-ODE forward using the joint denoiser $(D_\x,D_\z)$ to obtain $(\x_T,\z_T)$. We then run the PF-ODE backward using the tilted joint denoiser $(D_\x^r,D_\z^r)$. The similarity between the prompt and the decoded image increases monotonically with $\lambda$, both under CLIP, which defines the reward, and under SigLIP 2 \citep{tschannen2025siglip}, an image--text model trained separately from CLIP (\autoref{fig:prompt-editing}).

\subsection{Tilting a pretrained semantic prior}\label{sec:kandinsky-exp}
\begin{figure}[t]
    \centering
    \includegraphics[width=1\linewidth]{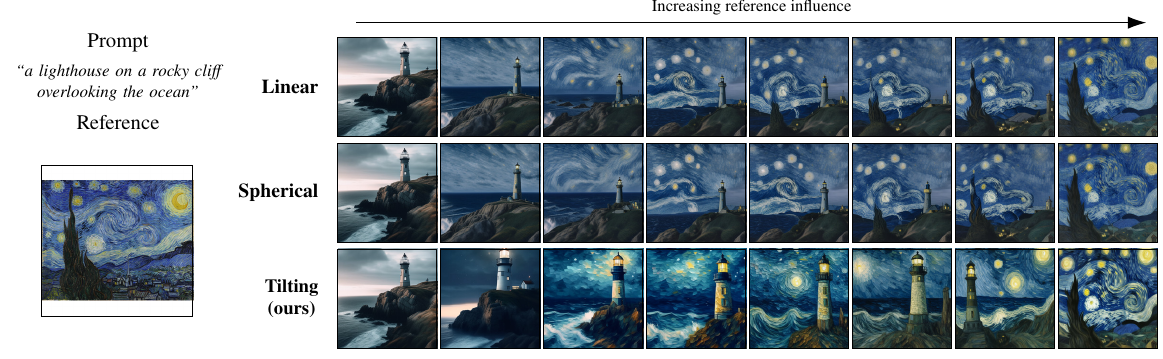}
    \caption{\textbf{Semantic tilting on a factorized model without retraining.} Kandinsky 2.2 samples from a text-conditional prior tilted toward the CLIP embedding of a reference image (Starry Night), compared with linear and spherical interpolation between the sampled and reference embeddings.}
    \label{fig:kandinsky-tilt}
\end{figure}
We next apply the factorized construction of \autoref{sec:embedding-only-new} to a pretrained model. Kandinsky 2.2 \citep{razzhigaev2023kandinsky} consists of a text-conditional diffusion prior $q(\z \mid \mathrm{prompt})$ and a separate decoder $p(\x \mid \z)$. We apply the shifted denoiser from \autoref{eq:general-denoiser-new} to the prior, which we sample without classifier-free guidance, and retrain neither component.

In \autoref{fig:kandinsky-tilt}, we tilt $q(\z \mid \mathrm{prompt})$ toward the embedding $\z^\star=\vphi(\x^\star)$ of a reference image $\x^\star$ with the reward $-\tfrac{\gamma}{2}\|\z-\z^\star\|^2$, and compare it to sampling from the untilted model and interpolating between the sampled embedding and $\z^\star$. In this example, tilting incorporates the reference style while retaining the lighthouse scene specified by the prompt. Unlike interpolation, the tilt has a well-defined target, the distribution that optimally balances proximity to the reference against the KL divergence from the text-conditional prior (\autoref{eq:tilt-objective}). Quantitatively, the tilted embeddings stay within the range of norms of untilted samples, unlike interpolated ones, but interpolation gives decoded images a better trade-off between style and prompt similarity (\appref{app:kandinsky}).

\section{Related work}\label{sec:related}
\paragraph{Reward alignment.} Generative models are commonly aligned with a reward either by fine-tuning \citep{pmlr-v202-lu23d,liu2025flowgrpo,domingo-enrich2025adjoint,bergmeister2026reinforceadjointmatchingscaling,potaptchik2026tilt} or by adding computation during sampling, such as gradient guidance \citep{chung2023diffusion} or sequential Monte Carlo \citep{skreta2025feynmankac,singhal2025a}. Fine-tuning is repeated for each reward, and typically for each reward strength, while inference-time methods increase the cost of every sample. Semantic noising amortizes a family of rewards into a single training run, after which any reward in the family can be applied at any strength without additional sampling cost.

\paragraph{Analytic tilting of diffusion models.} For linear and quadratic rewards, the denoiser of the tilted distribution can be expressed through the original denoiser at shifted inputs \citep{liu2024flow,moitra2026quadratic,mcdonald2026technical}, and posterior sampling from a noisy observation \citep{holderrieth2026glass} is a special case of the same identity (\appref{app:linear-tilt}). We show that, once applied to semantic representations, these identities give tilts by useful rewards such as proximity to a reference in semantic space and linear property predictors.

\paragraph{Joint and factorized models of data and representations.} Jointly diffusing data and their semantic representations improves sample quality and speeds up training \citep{kouzelis2025boosting,wu2025representation,kouzelis_coredi2026}. These models typically use a shared noise level for the data and the representation. We sample the two noise levels independently, as UniDiffuser \citep{pmlr-v202-bao23a} does for images and text, so that a joint model supports linear-quadratic semantic rewards. Factorized models such as unCLIP \citep{ramesh2022hierarchical} and Kandinsky 2.2 \citep{razzhigaev2023kandinsky}, which combine a semantic diffusion prior with a decoder, were designed for text-to-image generation and image variations, and support the same rewards without retraining.

\paragraph{Conditioning on noisy representations.} Conditioning a generator on a noise-corrupted input has been used to make cascaded diffusion models robust to errors of earlier stages \citep{ho2022cascaded} and, in Stable unCLIP \citep{Rombach_2022_CVPR,rombach2023stableunclip}, to generate variations of an image from its noisy CLIP embedding. By \autoref{eq:sem-noise-posterior}, a model trained in this way approximates exponential tilts and can therefore be used for reward control. We demonstrate this by training MolGPT on noisy CLAMP embeddings (\autoref{sec:molgpt-exp}) and by tilting Stable unCLIP without retraining (\appref{app:unclip}).

\section{Conclusion and limitations}
We proposed {\em semantic noising}, which allows a single generative model to be tilted by linear-quadratic rewards in semantic space, with objectives and strengths chosen after training, and demonstrated it in molecular and image generation with autoregressive and diffusion models.

Semantic noising also comes with limitations. First, the chosen representation determines the supported rewards. Rewards that are not linear-quadratic in $\vphi$ can only be approximated, with an error characterized by \autoref{prop:approx-reward}. Second, the approach is not black-box, since it requires a model conditioned on noisy representations, a joint diffusion model with independent noise levels, or a factorized model. For joint diffusion, independent noise levels give a slightly higher FID than shared noise levels (\appref{app:ablation}). Third, the tilting identities hold for the exact posteriors, whereas learned models only approximate them. Theoretically equivalent tilts can produce different results (\appref{app:gamma-ablation}). The MolGPT model gets roughly halfway to the exact tilt (\appref{app:exp-tilt-accuracy}), whereas the ImageNet model matches its exact target in LAION score and precision wherever we can estimate this target (\appref{app:realism}). Fourth, strong aesthetic tilts make the images less realistic (\appref{app:realism}). Finally, for Kandinsky 2.2 the tilted samples change abruptly with $\gamma$ for fixed noise, making the strength harder to tune (\appref{app:kandinsky}).

\subsection*{AI use statement}

In this work, we used generative AI tools to assist with writing experiment code, including implementing methods specified by the authors. We also used these tools to identify relevant literature, brainstorm, draft parts of the paper, and edit the text to improve readability. The authors reviewed all AI-assisted work, checked AI-generated code for correctness, and finalized all manuscript text.

We have {\em not} used generative AI tools to help develop theoretical models or conceptual frameworks, formulate mathematical claims, provide critical ingredients for proving mathematical claims, assist in writing proofs, propose or refine hypotheses, design or provide feedback on research methodology or experiments, support qualitative and thematic data analysis, or interpret results. The following tasks are not applicable to this work: generating synthetic datasets, assisting with translation, and cleaning and reformatting datasets.

We take responsibility for the final content of this work, including text, claims, and artifacts produced with the aid of generative AI.

\subsection*{Ethics statement}
This paper presents work that advances controllable generative models. While these advancements could enhance applications in creative and scientific domains, they also raise considerations around synthetic media generation and potential misuse. However, our contributions primarily aim at improving understanding and control in generative modeling, without introducing new ethical risks beyond those already associated with generative AI.

% \subsection*{Reproducibility statement}
% We provide experimental details in the appendix. Additionally, we upload code used for all our experiments as part of the submission.

% \subsubsection*{Author Contributions}
% If you'd like to, you may include  a section for author contributions as is done
% in many journals. This is optional and at the discretion of the authors.

\subsubsection*{Acknowledgments}
RK would like to thank Severi Rissanen, Najwa Laabid, Alison Pouplin, Maria Esteban, and Julien Martinelli for their feedback on the draft. This work was supported by the Finnish Center for Artificial Intelligence (FCAI) under Flagship R5
(award 15011052). The authors wish to acknowledge CSC -- IT Center for Science, Finland, for generous computational resources. PB acknowledges support from the Austrian Science Fund (FWF) 10.55776/DOC1345324.

\bibliography{refs}
\bibliographystyle{iclr2027_conference}

\appendix
\section{Exponential families}\label{app:exp-families}
A family of distributions $\lbrace p(\x \mid \vtheta)\rbrace_{\vtheta}$ parametrized by $\vtheta \in \R^k$ is a $k$-dimensional exponential family if its density can be written as \citep{fisher1934two,nielsen2009statistical}
\begin{equation}
    p(\x \mid \vtheta) \propto h(\x) \exp(\veta(\vtheta)\cdot T(\x)),
\end{equation}
where $h$ is the base measure, $\veta(\vtheta)$ is the \emph{natural parameter}, and $T(\x)$ is the \emph{sufficient statistic}. Since the distribution depends on $\vtheta$ only through $\veta(\vtheta)$, we can instead parametrize the family directly by its natural parameter:
\begin{equation}\label{eq:exp-fam}
    \pi(\x \mid \veta) \propto h(\x) \exp(\veta\cdot T(\x)).
\end{equation}
Then $p(\x \mid \vtheta) = \pi(\x \mid \veta(\vtheta))$.

\subsection{Tilting with rewards linear in $T$}
For a reward $r(\x) = \va\cdot T(\x)$, tilting $\pi$ gives
\begin{equation}\label{eq:exp-fam-tilting}
\begin{split}
    \pi^r(\x \mid \veta) &\propto h(\x) \exp(\veta\cdot T(\x)) \exp(\va\cdot T(\x)) = h(\x) \exp((\veta + \va)\cdot T(\x)) \\
    &  \propto \pi(\x \mid \veta + \va).
\end{split}
\end{equation}
Thus, tilting by a reward linear in the sufficient statistic stays within the family and shifts the natural parameter from $\veta$ to $\veta+\va$.

\subsection{Tilting with an approximate reward}\label{app:approx-reward}
Without the Gaussian assumption of \autoref{prop:approx-reward}, the divergence between the two tilts is determined by the variance of the approximation error $\delta(\x)=r(\x)-\hat r(\x)$ under intermediate tilts.
\begin{proposition}\label{prop:approx-reward-general}
Let $\hat r$ be an approximation of a reward $r$. For $s\in[0,1]$, let $r_s=(1-s)\,r+s\,\hat r$ interpolate between the two rewards. Then
\begin{equation}\label{eq:approx-reward-kl}
    \kl\bigl(q^{\hat r}\,\|\,q^{r}\bigr)=\int_0^1 s\,\Var_{q^{r_s}}[\delta]\,ds,
    \quad
    \kl\bigl(q^{r}\,\|\,q^{\hat r}\bigr)=\int_0^1 (1-s)\,\Var_{q^{r_s}}[\delta]\,ds.
\end{equation}
\end{proposition}
\paragraph{Proof of \autoref{prop:approx-reward-general}.} Since $r_s=r-s\,\delta$, the tilts $q^{r_s}$ form a one-dimensional exponential family with base measure $q(\x)\exp(r(\x))$, sufficient statistic $T(\x)=-\delta(\x)$, and natural parameter $s$. Let $A(s)=\log\E_{q}[\exp(r_s(\x))]$ be its log-partition function, so that $q^{r_s}(\x)=q(\x)\exp(r_s(\x)-A(s))$. Then
\begin{equation*}
    A'(s)=\E_{q^{r_s}}[T]=-\,\E_{q^{r_s}}[\delta],
    \qquad
    A''(s)=\Var_{q^{r_s}}[T]=\Var_{q^{r_s}}[\delta].
\end{equation*}
For $s,u\in[0,1]$, we have $\log q^{r_s}(\x)-\log q^{r_u}(\x)=(s-u)\,T(\x)-A(s)+A(u)$, and therefore
\begin{equation*}
    \kl\bigl(q^{r_s}\,\|\,q^{r_u}\bigr)=A(u)-A(s)-(u-s)\,A'(s),
\end{equation*}
which is the remainder of the first-order Taylor expansion of $A$ around $s$. Writing the remainder in integral form gives
\begin{align*}
    \kl\bigl(q^{r}\,\|\,q^{\hat r}\bigr)&=A(1)-A(0)-A'(0)=\int_0^1(1-s)\,A''(s)\,ds,\\
    \kl\bigl(q^{\hat r}\,\|\,q^{r}\bigr)&=A(0)-A(1)+A'(1)=\int_0^1 s\,A''(s)\,ds,
\end{align*}
and substituting $A''$ yields \autoref{eq:approx-reward-kl}.

\paragraph{Proof of \autoref{prop:approx-reward}.} Suppose that $r(\x)$ and $\hat r(\x)$ are jointly Gaussian for $\x\sim q$. For every $s$, the reward $r_s=(1-s)\,r+s\,\hat r$ is a linear combination of $r$ and $\hat r$ and is therefore Gaussian under $q$, with mean $\E_q[r]-s\,\E_q[\delta]$ and variance $\Var_q[r]-2s\Cov_q[r,\delta]+s^2\Var_q[\delta]$. A Gaussian random variable $Y$ satisfies $\E[\exp(Y)]=\exp(\E[Y]+\tfrac12\Var[Y])$, so the log-partition function from the proof of \autoref{prop:approx-reward-general} is
\begin{equation*}
    A(s)=\log\E_q[\exp(r_s(\x))]=\E_q[r]-s\,\E_q[\delta]+\tfrac12\bigl(\Var_q[r]-2s\Cov_q[r,\delta]+s^2\Var_q[\delta]\bigr).
\end{equation*}
Hence $\Var_{q^{r_s}}[\delta]=A''(s)=\Var_q[\delta]$ for every $s$, and \autoref{eq:approx-reward-kl} gives both divergences as $\tfrac{1}{2}\Var_q[\delta]$. Applying this result with $\hat r=0$, for which $\delta=r$ and $q^{\hat r}=q$, gives $\kl(q\,\|\,q^{r})=\tfrac{1}{2}\Var_q[r]$. Dividing the two yields $\kl(q^{\hat r}\,\|\,q^{r})/\kl(q\,\|\,q^{r})=\Var_q[\delta]/\Var_q[r]=1-R^2$.

\section{Direct conditional generation for autoregressive models}\label{app:algorithms}
\autoref{fig:sem-noise-alg} shows direct conditional generation (\autoref{sec:direct-cond-new}) for an autoregressive model of sequences $\x=(x_1,\dots,x_n)$, such as the MolGPT model of \autoref{sec:molgpt-exp}. Other generative models are trained in the same way, with their usual objective in place of the negative log-likelihood and $(\tilde\z,\tau_\z)$ as an additional conditioning input. If $\|\vphi(\x)\|$ is constant, $\tau_\z$ can be omitted from the conditioning input (\autoref{sec:direct-cond-new}), as in our MolGPT model.

\newcommand{\hlstep}[1]{\colorbox{blue!10}{#1}}
\newcommand{\algpanel}[1]{\hrule height 0.8pt\vspace{3pt}\textbf{#1}\vspace{3pt}\hrule\vspace{2pt}}
\begin{figure}[ht]
\begin{minipage}[t]{0.48\linewidth}
\algpanel{Training}
\begin{algorithmic}[1]
\Require data $q$, feature map $\vphi$, precision distribution $p(\tau)$
\Repeat
    \State $\x\sim q$
    \State \hlstep{$\tau_\z\sim p(\tau)$, \ $\eps\sim\mathcal{N}(\bm{0},\mI)$}
    \State \hlstep{$\tilde\z\gets\tau_\z\vphi(\x)+\sqrt{\tau_\z}\,\eps$}
    \State $\mathcal{L}\gets-\sum_{i}\log p_{\vtheta}(x_i\mid x_{<i},\hlstep{$\tilde\z,\tau_\z$})$
    \State $\vtheta\gets\vtheta-\eta\nabla_{\vtheta}\mathcal{L}$
\Until{converged}
\end{algorithmic}
\vspace{2pt}\hrule height 0.8pt
\end{minipage}\hfill
\begin{minipage}[t]{0.48\linewidth}
\algpanel{Sampling from the tilt by $r_{\va,\gamma}$}
\begin{algorithmic}[1]
\Require trained $p_{\vtheta}$, reward parameters $\va\in\R^k$, $\gamma>0$
\State \hlstep{$\tilde\z\gets\va$, \ $\tau_\z\gets\gamma$}
\State $i\gets0$
\Repeat
    \State $i\gets i+1$
    \State $x_i\sim p_{\vtheta}(\,\cdot\mid x_{<i},\hlstep{$\tilde\z,\tau_\z$})$
\Until{$x_i$ is the end token}
\State \Return $\x=(x_1,\dots,x_i)$
\end{algorithmic}
\vspace{2pt}\hrule height 0.8pt
\end{minipage}
\caption{Training and sampling with semantic noising for an autoregressive model. \protect\hlstep{Highlighted} steps differ from standard training and sampling.}\label{fig:sem-noise-alg}
\end{figure}

\section{Denoising posteriors form an exponential family}\label{app:exp-diffusion}
The denoising posterior $p_t(\x \mid \x_t)$ is the distribution of clean data given a noisy observation $\x_t$. By Bayes' rule,
\begin{equation}\label{eq:natural-posterior}
    p_t(\x\mid\x_t)
    \propto q(\x)\exp\left(
        \frac{\alpha_t\x_t}{\sigma_t^2}\cdot\x
        -\frac{\alpha_t^2}{2\sigma_t^2}\|\x\|^2
    \right).
\end{equation}
Therefore, for $\x\in\R^n$, the denoising posteriors form an $(n+1)$-dimensional exponential family with base measure $q$, sufficient statistic
\begin{equation}
    T(\x)=\left(\x,-\tfrac{1}{2}\|\x\|^2\right),
\end{equation}
and natural parameter \citep{karczewski2025spacetime}
\begin{equation}
    \veta(\x_t,t)
    =\left(\frac{\alpha_t \x_t}{\sigma_t^2},
    \frac{\alpha_t^2}{\sigma_t^2}\right)
    =(\tilde{\x},\tau).
\end{equation}
By expressing the denoising posterior directly in the natural parametrization, we have
\begin{equation}
    p_t(\x\mid\x_t)=\pi(\x\mid\veta(\x_t,t)).
\end{equation}
The denoiser in natural coordinates is
\begin{equation}
    D(\veta)=\int \x\,\pi(\x\mid\veta)d\x,
\end{equation}
and is related to the schedule-dependent denoiser by
\begin{equation}
    D_{(\alpha,\sigma)}(\x_t,t)
    =D(\veta(\x_t,t))
    =D\left(\frac{\alpha_t\x_t}{\sigma_t^2},
    \frac{\alpha_t^2}{\sigma_t^2}\right).
\end{equation}

\subsection{Linear-quadratic tilting in natural coordinates}\label{app:linear-tilt}\label{app:nat-quadr-tilt}
Consider the diffusion process obtained by applying the noising kernel to the tilted distribution $q^r\propto q\exp(r)$. The resulting denoising posterior is
\begin{equation}
    \pi^r(\x\mid\veta)
    \propto q^r(\x)\exp(\veta\cdot T(\x))
    \propto \pi(\x\mid\veta)\exp(r(\x)).
\end{equation}
The corresponding denoiser is
\begin{equation}\label{eq:tilted-denoiser-appendix}
    D^r(\veta)
    =\frac{\E[\x\exp(r(\x))\mid\veta]}
    {\E[\exp(r(\x))\mid\veta]},
\end{equation}
which is generally intractable. However, for
\begin{equation}
    r(\x)=\va\cdot\x-\tfrac{\gamma}{2}\|\x\|^2
    =(\va,\gamma)\cdot T(\x),
\end{equation}
the reward is linear in the sufficient statistic. Applying \autoref{eq:exp-fam-tilting} gives
\begin{equation}
    \pi^r(\x\mid(\tilde{\x},\tau))
    =\pi(\x\mid(\tilde{\x}+\va,\tau+\gamma)),
\end{equation}
and hence
\begin{equation}
    D^r(\tilde{\x},\tau)=D(\tilde{\x}+\va,\tau+\gamma).
\end{equation}

\paragraph{Posterior sampling as a special case.}
Let $\veta^\star=\veta(\x_t,t)$ be the natural parameter of a fixed noisy observation $\x_t$ at diffusion time $t$. By \autoref{eq:natural-posterior}, its posterior is $q$ tilted by $\veta^\star\cdot T(\x)$. Sampling from this posterior with the base diffusion model therefore uses
\begin{equation}\label{eq:glass}
    D_{\mathrm{posterior}}(\veta)=D(\veta+\veta^\star).
\end{equation}
This is the posterior-sampling formula of \citet{holderrieth2026glass} expressed in natural coordinates. Without this formula, posterior samples are typically drawn by integrating the reverse SDE from $(\x_t,t)$, since the PF-ODE started at $\x_t$ returns the same sample every time. \autoref{eq:glass} instead defines a diffusion model whose data distribution is the posterior, so independent posterior samples can be drawn from noise with the PF-ODE and its efficient solvers.

\subsection{Joint linear-quadratic tilting}\label{app:joint-quadratic-tilt}
Suppose that $\x$ and $\z$ are noised independently. The joint denoising posterior is an exponential family with natural parameter $\veta=(\tilde{\x},\tau_\x,\tilde{\z},\tau_\z)$ and sufficient statistic $T(\x,\z)=(\x,-\tfrac{1}{2}\|\x\|^2,\z,-\tfrac{1}{2}\|\z\|^2)$:
\begin{equation}
    \pi(\x, \z \mid \veta) \propto q(\x, \z) \exp\!\big(\veta\cdot T(\x, \z)\big).
\end{equation}
A linear-quadratic reward that depends only on the embedding is linear in the $\z$-components of $T$:
\begin{equation}
    r(\z)=\va\cdot\z-\tfrac{\gamma}{2}\|\z\|^2
    =(\bm{0},0,\va,\gamma)\cdot T(\x,\z).
\end{equation}
By \autoref{eq:exp-fam-tilting}, tilting shifts the natural parameter,
\begin{equation}
    \pi^r(\x,\z\mid(\tilde\x,\tau_\x,\tilde\z,\tau_\z))
    =\pi(\x,\z\mid(\tilde\x,\tau_\x,\tilde\z+\va,\tau_\z+\gamma)).
\end{equation}
Taking the expectations of $\x$ and $\z$ gives the tilted joint denoiser:
\begin{equation}
    (D_\x^r,D_\z^r)(\tilde\x,\tau_\x,\tilde\z,\tau_\z) =(D_\x,D_\z)(\tilde\x,\tau_\x,\tilde\z+\va,\tau_\z+\gamma).
\end{equation}

\section{MolGPT experimental details}\label{app:molgpt}
We train five MolGPT models for the experiments in \autoref{sec:molgpt-exp}: the noisy-embedding model, three direct property-conditioned models, and an unconditional reference model. Each is initialized from the released unconditional GuacaMol checkpoint \citep{molgpt}. The architecture has eight transformer layers, eight attention heads, width 256, and a 100-token context.

All five models are trained on the same 1,273,090 GuacaMol molecules \citep{guacamol}, in the same shuffled order and with the same schedule. We train for 100 epochs (331,600 updates) with batch size 384 and teacher-forced next-token prediction. We use AdamW with coefficients $(0.9,0.95)$, weight decay 0.1, and gradient clipping at 1. The learning rate warms up to $10^{-4}$ over 332 updates and then follows cosine decay, clipped at $10^{-5}$. The input token embeddings remain frozen.

\paragraph{Noisy-embedding conditioning.}
The frozen CLAMP molecular encoder \citep{pmlr-v202-seidl23a} produces 768-dimensional features from Morgan fingerprint counts, which we normalize to obtain $\vphi(\x)$. During training, we set $\tau_\z=0$ with probability 0.05 and otherwise sample $\sqrt{\tau_\z}\sim\mathrm{Uniform}[0.1,32]$, and then sample $\tilde{\z}$ from \autoref{eq:sem-noise-obs}. Since $\vphi(\x)$ is normalized, the model does not receive $\tau_\z$ (\autoref{sec:direct-cond-new}). We divide $\tilde{\z}$ by $32\sqrt{768+32^2}$, the root mean square norm of $\tilde{\z}$ at the largest $\tau_\z$, map it with a linear layer to three prefix tokens, and add a learned embedding to each of them.

\paragraph{Direct property conditioning.}
We train three independent models, one each for TPSA, logP, and SAS, following the original MolGPT property-conditioning setup. Each model maps its standardized property value with a learned linear layer to a prefix token.

\paragraph{Unconditional reference model.}
The unconditional model is trained with a constant prefix token. We use this model as a reference to estimate KL for the noisy-embedding and direct property-conditioned models.

\paragraph{Property directions.}
Molecular property values are not used to train the noisy-embedding generator. We fit each linear predictor by ridge regression on the training molecules, choosing the penalty on the validation molecules. For a predictor $\widehat R(\x)=\mathbf{w}\cdot\vphi(\x)+b$, we define the reward
\begin{equation}
    r_c(\x)
    =c\frac{\widehat R(\x)-\mu_{\widehat R}}{s_{\widehat R}}
    =\tilde{\z}\cdot\vphi(\x)+\mathrm{const},
    \qquad
    \tilde{\z}=\frac{c}{s_{\widehat R}}\mathbf{w}.
\end{equation}
At inference time, we condition on this $\tilde{\z}$, scaled as in training. At $c=0$, the conditioning input is $\tilde{\z}=\bm{0}$, which gives the untilted distribution.

\begin{table}[h]
    \centering
    \caption{\textbf{Linear property predictors for MolGPT.} $R^2$ is evaluated on 238,632 held-out molecules. The last column is the norm of $\tilde{\z}$ at $|c|=5$, divided by $32\sqrt{768+32^2}$ as in training. During training, this scaled norm ranges up to about 1.}
    \label{tab:molgpt-readouts}
    \begin{tabular}{lcc}
        \toprule
        Property & Held-out $R^2$ & Scaled $\|\tilde{\z}\|$ at $|c|=5$ \\
        \midrule
        Ring count & 0.780 & 0.615 \\
        Molecular weight & 0.723 & 0.520 \\
        Rotatable bonds & 0.585 & 0.761 \\
        TPSA & 0.750 & 0.648 \\
        logP & 0.778 & 0.565 \\
        SAS & 0.798 & 0.607 \\
        \bottomrule
    \end{tabular}
\end{table}

We chose ring count, molecular weight, and rotatable bonds from six additional candidate properties before inspecting their generation results. Ring count and molecular weight are predicted accurately from CLAMP, and the number of rotatable bonds measures molecular flexibility and is nearly uncorrelated with ring count.

\paragraph{Evaluation.}
For each condition, we generate 2,048 sequences from the initial token \texttt{C}, followed by at most 100 additional tokens, using temperature one and no top-$k$ restriction. RDKit \citep{rdkit} determines validity and evaluates the displayed properties.

For \autoref{fig:molgpt-reward-kl}, we estimate
\begin{equation}
    \kl(p\|p_{\mathrm{base}})
    =\E_{\x\sim p}\left[\log p(\x)-\log p_{\mathrm{base}}(\x)\right]
\end{equation}
for each conditional model $p$, using the exact autoregressive probability of each sampled sequence. Invalid sequences remain in the KL estimate, while property means are computed over valid molecules. For the noisy-embedding model, we use $c\in\{0,\pm0.5,\pm1,\pm1.5,\pm2,\pm2.5,\pm3,\pm4,\pm5\}$. For each direct property-conditioned model, the standardized target values range from $-1.5$ to $2.0$ in increments of 0.25.

\begin{figure}[tbh]
    \centering
    \includegraphics[width=\linewidth]{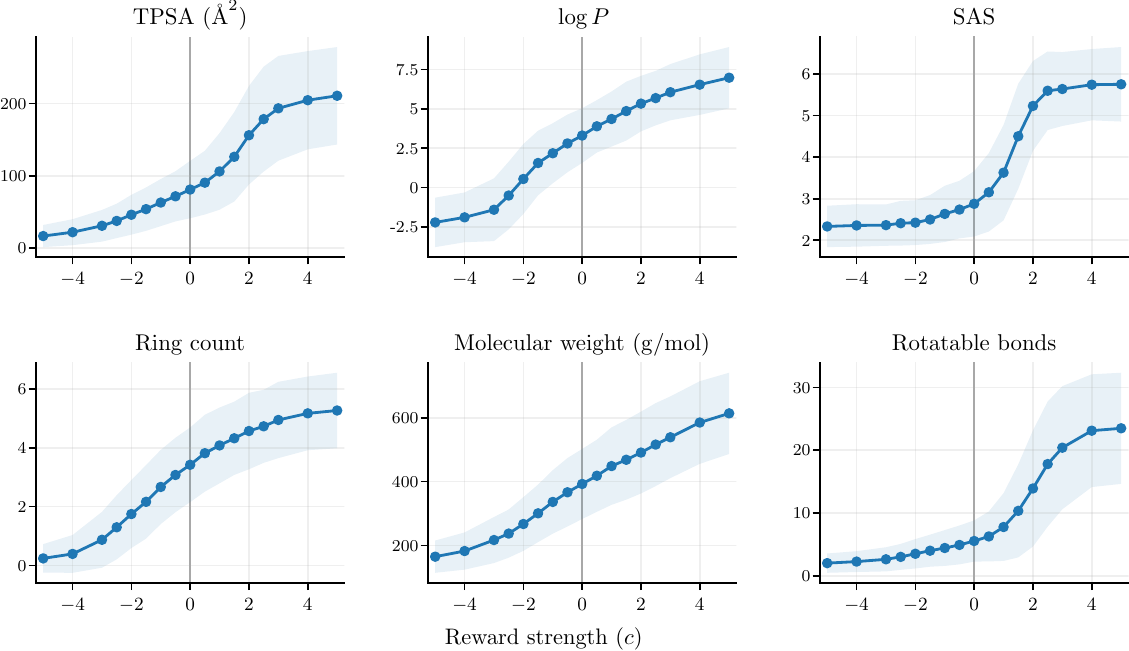}
    \caption{\textbf{Single-property control with one noisy-embedding generator.}
    In addition to the properties in \autoref{fig:molgpt-conditioning}, we show ring count, molecular weight, and rotatable bonds. Each panel varies the strength $c$ for one property and shows the mean RDKit property value. Shaded regions show one standard deviation among valid molecules.}
    \label{fig:molgpt-conditioning-full}
\end{figure}

For \autoref{fig:molgpt-conditioning-full}, the same grid of nonzero values of $c$ is used for each of the six properties, together with one shared $\tilde{\z}=\bm{0}$ distribution. About 97\% of the generated sequences are valid molecules, and over 99\% of these are unique within their condition.

\subsection{Exponential tilt accuracy}\label{app:exp-tilt-accuracy}
We measure the accuracy of the tilt with $\kl \left(p_{\vtheta}(\cdot \mid \tilde \z) \,\|\, q(\cdot \mid \tilde \z) \right)$, where
\begin{equation*}
    q(\x \mid \tilde \z) = \frac{q(\x)\exp(\tilde \z \cdot \vphi(\x) )}{Z(\tilde \z)}, \quad Z(\tilde \z) = \int q(\x) \exp(\tilde \z \cdot \vphi(\x) ) d\x.
\end{equation*}
The KL divergence can be approximated whenever we can evaluate the likelihoods of $p_{\vtheta}(\x \mid \tilde \z)$ and $q(\x)$ using:
\begin{align*}
    \kl \left(p_{\vtheta}(\cdot \mid \tilde \z) \,\|\, q(\cdot \mid \tilde \z) \right) &= \E_{\x \sim p_{\vtheta}(\x \mid \tilde \z)} \left[\log p_{\vtheta}(\x \mid \tilde \z) - \log q(\x \mid \tilde \z) \right] \\
    & = \E_{\x \sim p_{\vtheta}(\x \mid \tilde \z)} \left [ \rho(\x) \right] + \log \left( \E_{\x \sim q(\x)} \left[ \exp( \tilde \z \cdot \vphi(\x)) \right] \right) \\
    & = \E_{\x \sim p_{\vtheta}(\x \mid \tilde \z)} \left [ \rho(\x) \right] + \log \left( \E_{\x \sim p_{\vtheta}(\x \mid \tilde \z)} \left[ \exp(-\rho(\x)) \right] \right),
\end{align*}
where $\rho(\x) = \log p_{\vtheta}(\x \mid \tilde \z) - \log q(\x) - \tilde \z \cdot \vphi(\x)$. We use the last expression, which only requires samples from $p_{\vtheta}(\x \mid \tilde \z)$.

Since $q(\x)$ is not available, we use the model's own untilted distribution $p_{\vtheta}(\x \mid \tilde \z = \mathbf{0})$ in its place, so we measure how far $p_{\vtheta}(\cdot \mid \tilde \z)$ is from the exact tilt of $p_{\vtheta}(\cdot \mid \mathbf{0})$. Both distributions are restricted to valid SMILES strings, on which $\vphi$ is defined. For each of the six properties in \autoref{fig:molgpt-conditioning-full} and each $c\in\{\pm0.5,\pm1,\pm1.5,\pm2,\pm2.5,\pm3,\pm4,\pm5\}$, we generate 2{,}048 strings with $\tilde \z = c\mathbf{w}/s_{\widehat R}$ as in \autoref{sec:molgpt-exp}.

\begin{figure}[t]
    \centering
    \includegraphics[width=0.85\linewidth]{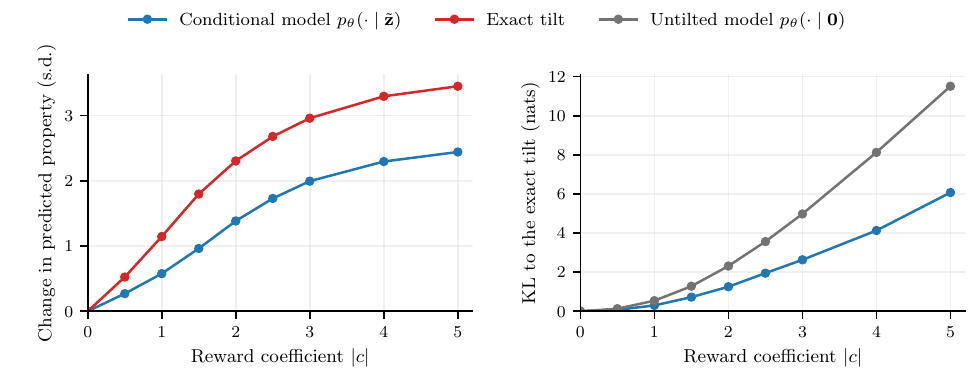}
    \caption{\textbf{The conditional model gets roughly halfway to the exact tilt.}
    Left: change in the mean predicted property $\widehat R$ in the requested direction, in units of $s_{\widehat R}$. Right: KL divergence from the exact tilt $q(\x\mid\tilde\z)\propto p_{\vtheta}(\x\mid\mathbf{0})\exp(\tilde\z\cdot\vphi(\x))$. All values are averaged over the six properties in \autoref{fig:molgpt-conditioning-full} and both signs of $c$.}
    \label{fig:molgpt-tilt-accuracy}
\end{figure}

A KL divergence is hard to interpret on its own, so we compare it with the divergence of the untilted model from the same target,
\begin{equation*}
    \kl \left(p_{\vtheta}(\cdot \mid \mathbf{0}) \,\|\, q(\cdot \mid \tilde \z) \right) = \log \left( \E_{\x \sim p_{\vtheta}(\x \mid \mathbf{0})} \left[ \exp( \tilde \z \cdot \vphi(\x)) \right] \right) - \E_{\x \sim p_{\vtheta}(\x \mid \mathbf{0})} \left[ \tilde \z \cdot \vphi(\x) \right],
\end{equation*}
which is the error we would make by ignoring the conditioning altogether. We also compare the change in the mean predicted property $\widehat R$ under the conditional model with the change under the exact tilt, which we estimate by reweighting samples from $p_{\vtheta}(\cdot \mid \mathbf{0})$ with $\exp(\tilde \z \cdot \vphi(\x))$. \autoref{fig:molgpt-tilt-accuracy} reports both comparisons averaged over the six properties and both signs of $c$.

By both measures, the conditional model gets roughly halfway to the exact tilt. For $|c|\leq1.5$, where the importance weights are still reliable, its change in the predicted property is 50--54\% of the change under the exact tilt, and for every $|c|$, its KL divergence from the exact tilt is 51--56\% of that of the untilted model. The conditional model therefore does not match the exact tilt, but it moves substantially toward it and reliably moves the properties in the requested direction (\autoref{fig:molgpt-conditioning-full}). A larger model or a richer conditioning mechanism than the linear map to prefix tokens may narrow the gap, which we have not tested.

All quantities except the change in the predicted property under the conditional model rely on importance weights, which concentrate on few samples as $|c|$ grows. Since the logarithm of a sample mean is biased downward, the KL divergences, as well as the change under the exact tilt, are likely underestimated when the effective sample size is small.

\paragraph{Best-of-$N$.} For $|c|\geq1.5$, Best-of-$N$ sampling from the untilted model, which keeps the molecule with the highest predicted property among $N$ samples, achieves a better reward--KL trade-off than our model, but at $N$ times its cost. From the 2{,}048 untilted samples, we find the smallest $N$ for which Best-of-$N$ matches the change in the predicted property of our model. Over the six properties and both signs of $c$, the median $N$ is 8 at $|c|=2$ and 58 at $|c|=5$. At these strengths, the KL divergence of Best-of-$N$ from the untilted model, which is at most $\log N-(N-1)/N$ \citep{beirami2025theoretical}, is below that of our model.

\paragraph{Coverage of the queries.} The norms of the queries $\tilde\z=c\mathbf{w}/s_{\widehat R}$ lie within the training range (\autoref{tab:molgpt-readouts}), but their directions do not. At these norms, a training input $\tilde\z=\tau_\z\vphi(\x)+\sqrt{\tau_\z}\bm{\epsilon}$ points roughly toward the embedding of its molecule, with a cosine similarity of 0.1 to 0.7, whereas every query has a cosine similarity below 0.04 with every training embedding. The model therefore extrapolates even for weak tilts, yet its KL divergence from the exact tilt stays at about half that of the untilted model (\autoref{fig:molgpt-tilt-accuracy}).

\subsection{Conditioning on exact rewards}\label{app:exact-rewards}
The CLAMP-based predictors approximate the properties with $R^2$ between 0.59 and 0.80 (\autoref{tab:molgpt-readouts}). To remove this approximation, we set $\vphi(\x)=(r_1(\x),\dots,r_7(\x))$, where $r_i$ are seven standardized RDKit properties, so that each property is an exact linear reward $r_i(\x)=\bm{e}_i\cdot\vphi(\x)$.

\paragraph{Sphere lifting.} As discussed in \autoref{sec:direct-cond-new}, a trained model cannot reliably represent linear rewards, which correspond to $\gamma=0$, unless $\|\vphi(\x)\|$ is constant. This is not a problem for the CLAMP embedding, which has unit norm, but the vector of properties $\vphi(\x)=(r_1(\x),\dots,r_7(\x))$ does not have constant norm. Normalizing it would give a representation of constant norm, but the individual properties are not linear functions of $\vphi(\x)/\|\vphi(\x)\|$, so they would no longer be exact linear rewards. In the extreme case of a single property, $\vphi(\x)=r(\x)$, the normalized representation $r(\x)/|r(\x)|$ is the sign of $r(\x)$, so a tilt could only change the relative weight of molecules with positive and negative values of the property. If $\|\z\|\leq R$, we instead use the representation
\begin{equation}\label{eq:sphere-lift}
    \w = \bigl(\z, \sqrt{R^2 - \|\z\|^2}\bigr),
\end{equation}
which lies on a sphere of radius $R$ in $\R^{k+1}$ and satisfies $(\va, 0)\cdot \w=\va \cdot \z$. A model trained with semantic noising of $\w$ and conditioned on $(\va,0)$ therefore targets the tilt by the linear reward $\va\cdot\z$ for every $\tau_\z>0$, so it does not need $\tau_\z$ as an input.

\paragraph{Setup.} The seven properties are TPSA, logP, SAS, ring count, molecular weight, rotatable bonds, and QED, each standardized on the training set. Since a few molecules with extreme values would otherwise determine the bound $R$ in \autoref{eq:sphere-lift}, we remove the training molecules whose standardized property vector has a norm above the 99th percentile, and the validation molecules above the same threshold. We then standardize the properties again on the remaining training molecules and set $R$ to 1.01 times the largest remaining norm.

We train two models on the filtered data with the backbone, initialization, frozen token embeddings, optimizer, learning-rate schedule, batch size, and number of epochs of \appref{app:molgpt}. The sphere-lifted model is trained with semantic noising of $\w$ and, like the CLAMP model, does not receive $\tau_\z$. A linear layer maps its 8-dimensional input to one prefix token. The unlifted model is trained with semantic noising of $\vphi(\x)$ and receives two prefix tokens, a linear projection of its 7-dimensional input and an encoding of $\tau_\z$ given by sinusoidal features of $\log\tau_\z$ followed by a two-layer MLP. As in \appref{app:molgpt}, each model divides its input by the root mean square norm of $\tilde\z$ at the largest $\tau_\z$. Both models are trained with $\sqrt{\tau_\z}\sim\mathrm{Uniform}[0.001,32]$, replaced by $\tau_\z=0.001$ with probability 0.05, and reach almost the same validation negative log-likelihood.

\begin{figure}[t]
    \centering
    \includegraphics[width=0.9\linewidth]{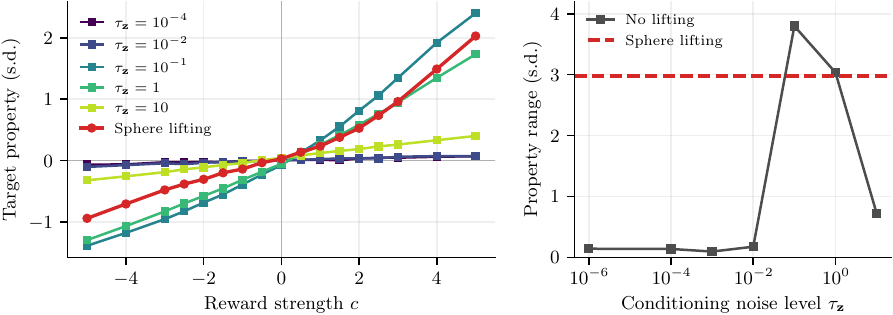}
    \caption{\textbf{Property control with exact rewards.}
    Left: mean target property against reward strength $c$, in standard deviations and averaged over seven properties, for the unlifted model at several $\tau_\z$ and for the sphere-lifted model. Right: change in the mean target property between $c=-5$ and $c=5$ as a function of $\tau_\z$.}
    \label{fig:molgpt-sphere-property-control}
\end{figure}

\begin{figure}[t]
    \centering
    \includegraphics[width=\linewidth]{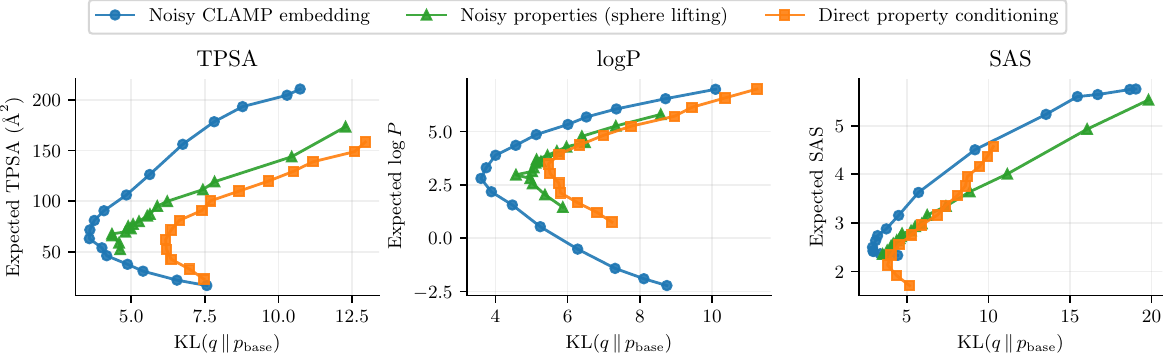}
    \caption{\textbf{Exact-reward conditioning does not improve the reward--KL trade-off over CLAMP-predicted rewards.}
    Same as \autoref{fig:molgpt-reward-kl}, with the sphere-lifted model added, which tilts by the exact properties.}
    \label{fig:molgpt-sphere-reward-kl}
\end{figure}

For property $i$ and $c\in\{\pm0.5,\pm1,\pm1.5,\pm2,\pm2.5,\pm3,\pm4,\pm5\}$, we condition the sphere-lifted model on $(c\bm{e}_i,0)$ and the unlifted model on $\tilde\z=c\bm{e}_i$ with $\tau_\z\in\{10^{-6},10^{-4},10^{-3},10^{-2},0.1,1,10\}$. We generate 2,048 sequences per condition as in \appref{app:molgpt} and estimate the KL divergence from the same unconditional reference model as in \autoref{fig:molgpt-reward-kl}. At $|c|=5$, about 95\% of the generated sequences are valid molecules for the sphere-lifted model and for the unlifted model at $\tau_\z\in\{0.1,1\}$. \autoref{fig:molgpt-sphere-property-control} shows the response curves, and \autoref{fig:molgpt-sphere-reward-kl} adds the sphere-lifted model to the reward--KL comparison of \autoref{fig:molgpt-reward-kl}.

\paragraph{Results.} For $\tau_\z\leq10^{-2}$, conditioning the unlifted model barely changes the properties (\autoref{fig:molgpt-sphere-property-control}). A larger $\tau_\z$ gives strong control, but the target then also includes the quadratic term $-\tfrac{\tau_\z}{2}\|\vphi(\x)\|^2$, which dominates at $\tau_\z=10$. The sphere-lifted model gives strong control without choosing $\tau_\z$. However, exact rewards do not improve the reward--KL trade-off over the CLAMP model (\autoref{fig:molgpt-sphere-reward-kl}). Even without tilting, the sphere-lifted model is further from the unconditional reference model than the CLAMP model (KL 5.26 vs 3.75 nats).

\section{Joint diffusion model ablation study}\label{app:ablation}

REG \citep{wu2025representation} jointly diffuses image latent tokens and an extra embedding token extracted from a pretrained image encoder. It also uses REPA \citep{yu2025representation} to align intermediate denoiser patch representations with those of the same encoder during training. The original study reports that DINOv2 \citep{oquab2024dinov} outperforms CLIP \citep{pmlr-v139-radford21a} for generation quality when used as both the REPA target and the generated embedding. Our downstream applications nevertheless require a CLIP coordinate, since text-based editing uses CLIP's text embedding, and the LAION aesthetic predictor \citep{schuhmann2022laionb} is a linear projection of its image embedding. We therefore study how to add a CLIP coordinate without sacrificing generation quality.

We reproduced REG at B/2 scale as a calibration baseline (FID 15.50 at 400K training steps compared with the reported 15.22) and conducted all ablations at this scale before scaling the selected recipe to XL/2.

\paragraph{Metrics.} We report FID, computed from 50K class-conditioned samples without classifier-free guidance, and CLIP consistency. To measure CLIP consistency, we fix a clean CLIP embedding as the conditioning input ($t_\z=0$), generate an image with the usual image and DINO noise schedule, re-encode it with CLIP, and compute its cosine similarity with the conditioning embedding, averaged over 1{,}024 samples.

\subsection{Representation choice}\label{app:representation}

Our first model replaced DINOv2 with CLIP throughout, using CLIP patch tokens as REPA targets and a normalized CLIP global embedding as the generated coordinate. This confirmed REG's finding, with FID 43.27 at 100K steps compared with 35.39 for our REG reproduction. The REPA patch targets are used only during training, whereas the global coordinate is jointly diffused with the image and generated at inference time, so we varied the two choices separately, with shared noise levels. At 100K steps, switching the REPA targets to DINOv2 patches improves FID to 38.31, and further replacing the CLIP global coordinate with the DINOv2 CLS token improves it to 35.30, about the FID of REG.

\paragraph{Dual DINO+CLIP coordinates.} Rather than choosing between DINO and CLIP, we keep the DINO CLS token as the primary generated coordinate and add a second generated token for the normalized CLIP embedding of the image. The CLIP token has its own input projection, layer normalization and output head, which add 0.85\% parameters, and its velocity loss has weight 0.03. This dual model has the best B/2 FID, 33.05 at 100K steps and 13.75 at 400K steps, compared with 15.50 for REG. Sampling the DINO noise level independently as well would allow tilting in both semantic spaces, which we did not pursue.

\subsection{Noise schedule}\label{app:noise-schedule}

During standard generation, all coordinates use the same noise level, and training the dual model with shared noise gives the best FID. Semantic tilting, however, changes the CLIP precision without changing the image precision (\autoref{eq:joint-tilted-denoiser-new}), and a shared-noise model is never trained on such pairs of noise levels, for example a clean CLIP embedding with a noisy image.

\begin{figure}[h]
    \centering
    \includegraphics[width=0.8\linewidth]{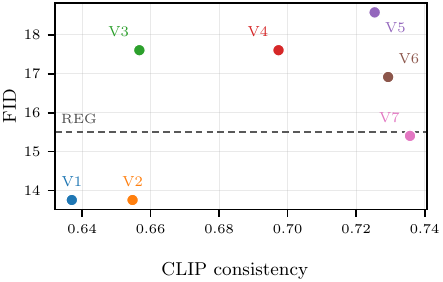}
    \caption{\textbf{Generation quality against CLIP consistency at B/2 scale.} FID and CLIP consistency at 400K training steps, where a lower FID and a higher consistency are better. The dashed line marks REG.}
    \label{fig:fid-consistency}
\end{figure}

We compare seven variants at 400K steps (\autoref{fig:fid-consistency}). We use the shorthand $(t_{\mathrm{img}}, t_{\mathrm{DINO}}, t_\z)$ for the noise levels of the image latent, DINO, and CLIP coordinates, respectively, where $t$ denotes the shared diffusion time and $U_1, U_2, U_3 \sim \mathrm{Uniform}[0, T]$ are independent draws.

\begin{description}
\item[V1] DINO+CLIP, all noise levels shared $(t, t, t)$, one shared timestep MLP.

\item[V2] The V1 model, evaluated with the timestep input of the clean CLIP embedding set to $t_{\mathrm{img}}$ rather than $0$.

\item[V3] CLIP only $(t, t)$, shared noise and shared timestep MLP. This is the model with DINOv2 patch targets and a CLIP global coordinate from \autoref{app:representation}.

\item[V4] The V3 model, evaluated as V2.

\item[V5] DINO+CLIP, all noise levels independent $(U_1, U_2, U_3)$, one shared timestep MLP.

\item[V6] As V5, with three separate timestep MLPs.

\item[V7] DINO+CLIP, image and DINO share the noise level and CLIP is independent $(U_1, U_1, U_2)$, with a separate CLIP timestep MLP.
\end{description}

Models trained with independent CLIP noise (V5--V7) achieve substantially higher CLIP consistency than the shared-noise models. Among them, separate timestep MLPs improve FID (V6 vs.\ V5), and sharing the noise level of the image and DINO coordinates improves it further to about the FID of REG (V7), while the CLIP consistency of V5 to V7 is similar. We trained each configuration once and selected V7 for scaling to XL. Two further variants, in which the image and DINO coordinates share the noise level $U_1$ and the CLIP noise level is either $\min(U_1, U_2)$ or a uniformly random fraction of $U_1$, had a worse FID than V7 at 100K steps and were stopped.

\subsection{Scaling to XL and practical overhead}\label{app:scaling}

We scaled the selected V7 recipe to SiT-XL/2 (688M parameters). Without caching, the dominant additional cost is the CLIP encoder forward pass used to compute training targets. Since the training crops are fixed, we precomputed and cached all CLIP global embeddings, which took about three minutes on four GH200 GPUs. Relative to REG, the model has 3.40M (0.50\%) more parameters, costs 0.26\% more per training update with cached CLIP targets, and 0.78\% more per denoiser evaluation.

At 800K training steps, the XL model reaches FID 2.66 and CLIP consistency 0.82 without classifier-free guidance. At 2.4M steps, FID improves to 2.32 and CLIP consistency to 0.833, while REG reports FID 2.2.

\section{Quadratic coefficient ablation}\label{app:gamma-ablation}
\begin{figure}[h]
    \centering
    \includegraphics[width=\linewidth]{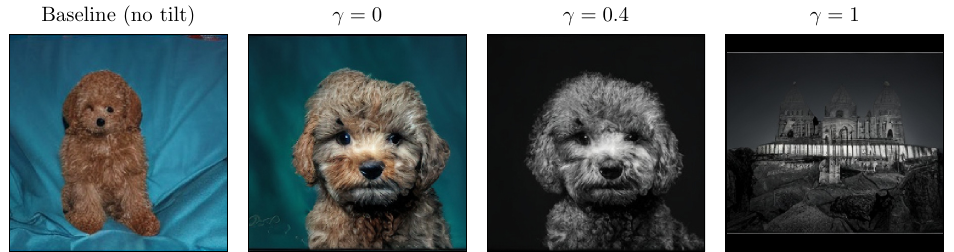}
    \caption{\textbf{The learned model depends on $\gamma$ although the target distributions are the same.} Aesthetic tilting at $\lambda=3$ with normalized CLIP embeddings and different values of $\gamma$.}
    \label{fig:gamma-ablation}
\end{figure}

For normalized CLIP embeddings, the quadratic term of the aesthetic reward is constant, so $\gamma$ does not change the target distribution (\autoref{eq:sem-noise-posterior}). It does change the inputs at which the learned denoiser is evaluated, and \autoref{fig:gamma-ablation} shows that different values of $\gamma$ can give very different samples at the same strength, so $\gamma$ is a hyperparameter of our method. The choice arises only for models that receive $\tau_\z$ as an input. For normalized embeddings, this input is not needed (\autoref{sec:direct-cond-new}), and our MolGPT model is trained without it. We chose $\gamma=0.01$ by inspecting samples and use it for all strengths. A more principled choice would evaluate the denoiser at the equivalent input with the highest density under the training distribution, which we leave for future work.

\begin{figure}[h]
    \centering
    \includegraphics[width=0.85\linewidth]{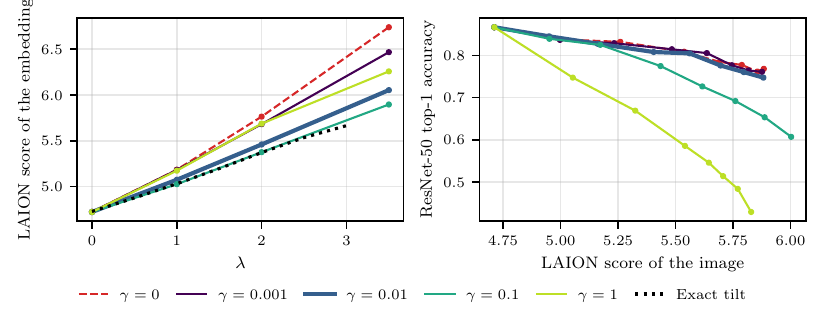}
    \caption{\textbf{$\gamma=0.1$ follows the exact tilt, but $\gamma\leq0.01$ keeps the class better at strong tilts.} Five of the eight values of $\gamma$ on 24 ImageNet classes. Left: LAION score of the generated CLIP embedding, with the exact tilt estimated up to $\lambda=3$. Right: ResNet-50 top-1 accuracy against the LAION score of the decoded image, for $\lambda$ up to 15.}
    \label{fig:gamma-sweep}
\end{figure}

To measure the effect of $\gamma$, we ran our method with eight values of $\gamma$ between 0 and 1 on 24 ImageNet classes disjoint from those of \appref{app:steering}, with the same sampler and number of outputs, and estimated the exact tilt from 1,024 untilted samples per class as in \appref{app:realism}. For $\gamma=0$, the shift $(\sigma_t^2/\alpha_t)\va$ in standard coordinates diverges as $\alpha_t\to0$ (\autoref{sec:lq-tilting-new}), so we use the untilted denoiser at the first sampling step. Every value of $\gamma$ raises the aesthetic score, but by a different amount (\autoref{fig:gamma-sweep}, left). The LAION score of the generated embedding follows the exact tilt for $\gamma$ between 0.03 and 0.1 and overshoots it increasingly as $\gamma$ moves away from this range. However, $\gamma$ above 0.01 generates fewer images of the conditioning class, which we measure by the top-1 accuracy of a ResNet-50 ImageNet classifier \citep{He_2016_CVPR} at the same LAION score of the image (\autoref{fig:gamma-sweep}, right). Among the values with the highest accuracy, $\gamma=0.01$ follows the exact tilt most closely. Its generated embeddings still overshoot the exact tilt, but the decoded images match it in LAION score and precision up to $\lambda=3.5$ (\appref{app:realism}).

\section{Comparison with inference-time steering}\label{app:steering}

\paragraph{Setup.} We use the XL joint model after 2.4M training steps and the sampler of \autoref{sec:joint-exp} (250 Euler--Maruyama steps without classifier-free guidance). Each method produces 64 outputs for each of 24 classes evenly spaced over the ImageNet labels, which gives 1,536 outputs per configuration. Particle $j$ of output $o$ starts from the same initial noise in every method, and our runs at different strengths share their Brownian increments. All methods target $p(\x\mid c)\exp(\lambda S/\sigma_S)$, where $S$ is the LAION aesthetic score and $\sigma_S=0.486$ is its standard deviation over untilted samples, and we call $S/\sigma_S$ the steering reward. Our method computes $S$ from the generated CLIP embedding, as in the aesthetic tilt of \autoref{sec:joint-exp}, and the baselines compute it from the CLIP embedding of the decoded image.

\paragraph{Baselines.} Best-of-$N$ returns the sample with the highest reward among $N$ independent samples. DPS \citep{chung2023diffusion} adds $\lambda$ times the gradient of $S(D_\x)/\sigma_S$, the steering reward of the decoded denoiser output, to the score at every step after the first, which approximates the gradient of $\log\E[\exp(\lambda S/\sigma_S)\mid\tilde\x,\tau_\x,\tilde\z,\tau_\z]$. The gradient is taken with respect to the joint noisy state, through CLIP, the VAE decoder and the denoiser. FKS \citep{singhal2025a} propagates $K$ particles per output and resamples them multinomially at steps 12, 24, \dots, 72 of 250, which corresponds to the 5\% to 30\% window of the reference implementation. We use the difference potential $\exp(\lambda(r_k-r_{k-1}))$, where $r_k$ is the look-ahead reward at the $k$-th resampling step, and divide the final weights by the potentials along each particle's ancestry. We report the final particle with the highest reward, which favors FKS over a draw from the final weights. The look-ahead reward is computed in one of three ways.
\begin{itemize}
    \item The mean-image look-ahead scores the decoded denoiser output $D_\x$, as in the reference implementation.
    \item The posterior look-ahead scores $M$ samples from $p(\x_0\mid\x_t)$, drawn with the posterior denoiser of \autoref{eq:glass} and an ODE sampler with $\mathrm{round}(32t)$ Heun steps. For $M>1$, the samples are combined as $\frac{1}{\lambda}\log\frac{1}{M}\sum_{m=1}^M\exp(\lambda r(\x_0^{(m)}))$, following the many-sample look-ahead of \citet{singhal2025a}.
    \item The semantic look-ahead scores the denoised CLIP embedding $D_\z$ and needs no decoding. It requires a model that generates the CLIP embedding, such as our joint model, and is not available for diffusion models of images alone. Since the LAION score $S$ is linear in the CLIP embedding, the expected score of the generated CLIP embedding given the current noisy state equals the score of the denoised CLIP embedding, $\E[S(\z)\mid\tilde\x,\tau_\x,\tilde\z,\tau_\z]=S(D_\z(\tilde\x,\tau_\x,\tilde\z,\tau_\z))$.
\end{itemize}

\paragraph{Strengths.} We evaluate our method at $\lambda\in\{0,0.5,\dots,3.5,5,7,10,12,15,20,25,40,60,100\}$ and DPS at $\lambda\in\{1,2,3.5,5,7,10,15,25,40\}$. Best-of-$N$ uses $N\in\{1,2,4,8,16\}$. FKS uses $K\in\{4,16\}$ particles, with $\lambda\in\{0.5,1,2,4,10\}$ for the mean-image look-ahead and $\lambda\in\{1,2,4,10,20,40\}$ for the semantic look-ahead. With $M=4$ and $K=16$, the posterior look-ahead costs 32 GPU-seconds per output, so we take it from an earlier run with 8 outputs per class, at $\lambda=4$, which gave a higher steering reward than $\lambda=1$.

\paragraph{Metrics and cost.} For every method, the reported LAION score is computed from the CLIP embedding of the decoded image. We compute MUSIQ-AVA \citep{ke2021musiq} and NIMA \citep{talebi2018nima} with IQA-PyTorch \citep{chen2022iqapytorch}. Both are trained on AVA ratings \citep{murray2012ava}, as is the LAION predictor we use (together with LAION-Logos ratings), so the three scores measure related notions of aesthetic quality with different architectures and inputs. Diversity is one minus the mean pairwise cosine similarity between the DINOv2 embeddings of the 64 outputs of a class, averaged over classes. GPU-seconds are the measured sampling time per output on GH200 GPUs at batch size 256. They include the look-aheads and exclude the final decoding and scoring of each particle, which takes about 6\,ms. An untilted sample and our method take 0.31 GPU-seconds per output, DPS 3.8, and FKS with 16 particles 4.8 with the semantic and 5.4 with the mean-image look-ahead.

\paragraph{Precision.} We compute the precision \citep{kynkaanniemi2019improved} of generated images $\hat\x_1,\dots,\hat\x_M$ against real ImageNet training images $\x_1,\dots,\x_n$ of the same class,
\begin{equation}\label{eq:precision}
    \mathrm{precision}=\frac{1}{M}\sum_{j=1}^{M}\mathbbm{1}\Bigl[\exists\, i\colon\ \|f(\hat\x_j)-f(\x_i)\|\le d_5(\x_i)\Bigr],
\end{equation}
where $f$ denotes DINOv2 features \citep{oquab2024dinov} and $d_5(\x_i)$ is the distance from $f(\x_i)$ to its fifth nearest neighbor among the features of the real images. A generated image thus counts toward precision if it lies in the neighborhood of at least one real image. Held-out real images (50 per class) reach a precision of 0.87.

\paragraph{Results.} \autoref{tab:steering} compares the methods at a LAION score of about 5.4, which our method reaches at $\lambda=3.5$, and of about 5.8, which it reaches at $\lambda=12$. At the same LAION score, our method is higher in MUSIQ and NIMA than every baseline, over the whole range of LAION scores that our method reaches. FKS keeps a higher precision than our method, whereas DPS has a similar precision and a higher diversity. Best-of-$N$ and FKS improve with the number of particles at a cost that grows linearly with it, whereas our method improves with $\lambda$ at the cost of a single sample. FKS resamples particles by an estimate of their final reward, and its results depend on how well this estimate predicts the final reward. The mean-image look-ahead scores the decoded denoised image, which early in sampling averages over many possible final images. Its score appears to be a poor estimate of the final score, since larger strengths make the results worse (with 16 particles, LAION score 5.34 at $\lambda=0.5$ and 5.17 at $\lambda=10$). The posterior look-ahead with four samples gives the best results among the FKS variants that score images, at about 6.5 times the cost of Best-of-16. The semantic look-ahead computes the expected score of the generated CLIP embedding in closed form (\autoref{eq:semantic-fks}), and this score is correlated with that of the decoded image (\appref{app:realism}). It needs no additional denoiser evaluations and gives the strongest FKS results. For $\lambda\ge7$, DPS reaches LAION scores above the largest that our method reaches (6.0 at $\lambda=100$), while its precision keeps decreasing, to 0.27 at $\lambda=40$.

\begin{table}[h]
    \centering
    \caption{\textbf{Steering at a moderate and a strong LAION score.} The methods in the middle group reach a LAION score of about 5.4, and those in the bottom group about 5.8. FKS uses 16 particles, and its posterior look-ahead averages four posterior samples. Standard errors (grey) are over outputs, and over classes for diversity. $^\ast$From an earlier run with 8 outputs per class.}
    \label{tab:steering}
    \footnotesize
    \begin{tabular}{lccccc}
        \toprule
        Method & LAION & MUSIQ & NIMA & Precision & Diversity \\
        \midrule
        Base model & 4.71\se{0.01} & 4.18\se{0.01} & 4.38\se{0.01} & 0.62\se{0.01} & 0.55\se{0.03} \\
        \faintrule
        Best-of-16 & 5.41\se{0.01} & 4.63\se{0.01} & 4.93\se{0.01} & 0.60\se{0.01} & 0.50\se{0.03} \\
        DPS, $\lambda=3.5$ & 5.47\se{0.01} & 4.36\se{0.01} & 4.69\se{0.01} & 0.56\se{0.01} & 0.57\se{0.02} \\
        FKS-16, mean image, $\lambda=0.5$ & 5.34\se{0.01} & 4.56\se{0.01} & 4.84\se{0.01} & 0.64\se{0.01} & 0.50\se{0.03} \\
        FKS-16, posterior, $\lambda=4$$^\ast$ & 5.56\se{0.03} & 4.69\se{0.04} & 5.02\se{0.04} & 0.63\se{0.03} & 0.49\se{0.04} \\
        FKS-16, semantic, $\lambda=1$ & 5.46\se{0.01} & 4.66\se{0.01} & 4.97\se{0.01} & 0.60\se{0.01} & 0.49\se{0.03} \\
        Ours, $\lambda=3.5$ & 5.37\se{0.01} & 4.73\se{0.01} & 5.09\se{0.01} & 0.59\se{0.01} & 0.52\se{0.03} \\
        \faintrule
        DPS, $\lambda=5$ & 5.83\se{0.01} & 4.41\se{0.01} & 4.81\se{0.01} & 0.51\se{0.01} & 0.59\se{0.02} \\
        FKS-16, semantic, $\lambda=10$ & 5.79\se{0.01} & 4.92\se{0.01} & 5.30\se{0.01} & 0.58\se{0.01} & 0.48\se{0.04} \\
        Ours, $\lambda=12$ & 5.81\se{0.01} & 5.09\se{0.01} & 5.64\se{0.01} & 0.51\se{0.01} & 0.54\se{0.03} \\
        \bottomrule
    \end{tabular}
\end{table}

\subsection{Exact tilts and precision}\label{app:realism}
\begin{figure}[h]
    \centering
    \includegraphics[width=0.5\linewidth]{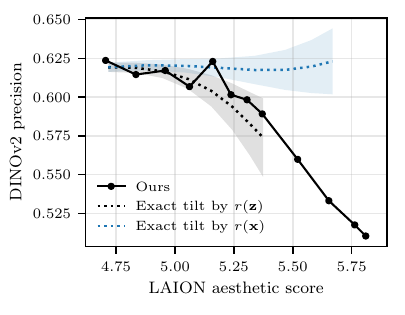}
    \caption{\textbf{Our method follows its exact target up to $\lambda=3.5$.} Precision against the LAION score. Dotted lines show the exact tilts by $r(\z)$, targeted by our method, and by $r(\x)$, targeted by the baselines, estimated by reweighting 1,024 untilted samples per class, with shaded bootstrap standard errors.}
    \label{fig:steering-exact}
\end{figure}

\paragraph{Exact tilts.} Our method and the baselines use the same learned model $p_{\vtheta}(\x,\z)$ but target different tilts of it,
\begin{equation}\label{eq:rz-rx}
    p_{\vtheta}^{r(\z)}(\x,\z)\propto p_{\vtheta}(\x,\z)\exp\bigl(\lambda S(\z)/\sigma_S\bigr),
    \qquad
    p_{\vtheta}^{r(\x)}(\x,\z)\propto p_{\vtheta}(\x,\z)\exp\bigl(\lambda S(\vphi(\x))/\sigma_S\bigr),
\end{equation}
where we omit the class label. Our method targets the tilt by the score of the generated CLIP embedding, $r(\z)$, whereas the baselines target the tilt by the score of the decoded image, $r(\x)$. For the data distribution, $\z=\vphi(\x)$ and the two tilts coincide. We estimate both exact tilts by self-normalized reweighting of 1,024 untilted samples per class, from which we also draw the Best-of-$N$ outputs. The effective sample size decreases with $\lambda$ and averages 18 per class at $\lambda=3.5$, so we do not estimate stronger tilts. For the learned model, the scores of the generated embedding and of the decoded image have a within-class correlation of 0.70. At equal $\lambda$, the $r(\z)$ tilt therefore raises the LAION score of the image less than the $r(\x)$ tilt, but at the same LAION score its precision is only $0.04\pm0.02$ lower. Up to $\lambda=3.5$, our method matches its exact target in both LAION score and precision (\autoref{fig:steering-exact}). At this strength, which we use in the main text, it also matches Best-of-16 in LAION score and precision.

\paragraph{Strong tilts.} Beyond $\lambda=3.5$, the precision of our method keeps decreasing, and the decoded images follow the generated CLIP embeddings less closely. At $\lambda=10$, the generated CLIP embeddings have a mean LAION score about nine within-class standard deviations above that of untilted samples. The cosine similarity between the generated CLIP embedding $\z$ and the CLIP embedding $\vphi(\x)$ of the decoded image is 0.82 without tilting and 0.81 at $\lambda=3.5$, but falls to 0.72 at $\lambda=10$. Even without tilting, samples with a lower similarity have a lower precision. When we split the untilted samples of each class into four equal groups by this similarity, the precision rises from 0.57 in the group with the lowest similarity to 0.68 in the group with the highest. FKS selects among untilted particles and does not shift the inputs of the denoiser, and for its outputs this similarity is 0.84.

\subsection{Self-recurrence}\label{app:recurrence}
\begin{figure}[h]
    \centering
    \includegraphics[width=\linewidth]{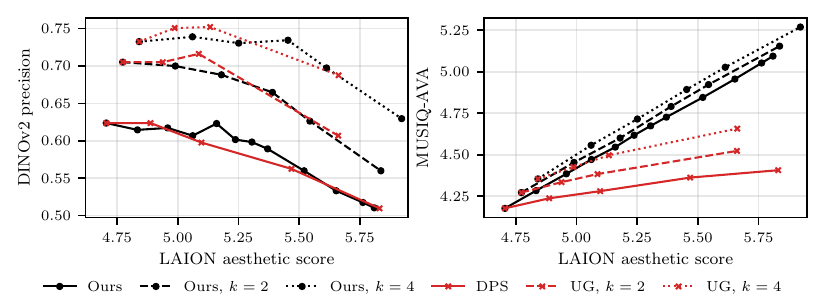}
    \caption{\textbf{Self-recurrence raises the precision of DPS and of our method.} Precision (left) and MUSIQ score (right) against the LAION score, without self-recurrence and with $k=2$ and $k=4$ repetitions per step. DPS with self-recurrence is universal guidance (UG).}
    \label{fig:steering-recurrence}
\end{figure}

Universal guidance \citep{bansal2024universal} adds self-recurrence to DPS. Each sampling step after the first is taken $k$ times, and after each of the first $k-1$ repetitions the state is renoised back to the current noise level, so that sampling costs $k$ times as much. We use $k\in\{2,4\}$ with $\lambda\in\{1,2,5,10,25\}$ and, as in the Stable Diffusion experiments of \citet{bansal2024universal}, without backward guidance. Self-recurrence changes the sampler but not the target, so we also apply it to our method, with $\lambda\in\{0,1,2,3.5,5,10,25\}$. Without tilting, it raises the precision from 0.62 to 0.71 for $k=2$ and to 0.73 for $k=4$, and lowers the diversity from 0.55 to 0.50 and 0.49. At the same LAION score, self-recurrence raises the precision of DPS and of our method by a similar amount (\autoref{fig:steering-recurrence}). With the same $k$, universal guidance and our method differ by at most 0.02 in precision, our method is higher in MUSIQ and NIMA, and universal guidance is slightly more diverse. Our method with self-recurrence takes 0.61 GPU-seconds per output for $k=2$ and 1.2 for $k=4$, about 13 times less than universal guidance with the same $k$. Since self-recurrence raises the precision even without tilting, it could also be combined with Best-of-$N$ and FKS, which we did not evaluate.

\section{Tilting with Stable unCLIP}\label{app:unclip}
Stable unCLIP is a pretrained model trained to condition on a noisy semantic representation, so it supports direct conditional generation (\autoref{sec:direct-cond-new}) without retraining. Unlike in our Kandinsky 2.2 experiments (\autoref{sec:kandinsky-exp}), which shift the inputs of a diffusion prior over embeddings and decode the sampled embedding with a separate decoder, the reward is specified through the conditioning input alone.

\paragraph{Model.} Stable unCLIP \citep{rombach2023stableunclip} is Stable Diffusion 2.1 \citep{Rombach_2022_CVPR} fine-tuned to generate images conditioned on a text prompt and a noisy CLIP image embedding, following unCLIP \citep{ramesh2022hierarchical}. We use the \texttt{small} variant, which conditions on the OpenAI CLIP ViT-L/14 image embedding \citep{pmlr-v139-radford21a}. During training, the CLIP embedding of each training image is standardized with fixed per-dimension statistics to $\vu=\vphi(\x)$ and noised as $\vu_s=\alpha_s\vu+\sigma_s\eps$, with $\alpha_s^2+\sigma_s^2=1$ and a random level $s\in\{0,\dots,999\}$. The noisy embedding $\vu_s$ and its level $s$ enter the model only as conditioning inputs, together with the prompt. The diffusion process runs over the image latent, whose noise is independent of $\eps$, so the model learns to denoise the image and never denoises the embedding. After rescaling, the conditioning input $\tilde\z=(\alpha_s/\sigma_s^2)\vu_s$ is semantic noising (\autoref{eq:sem-noise-obs}) of $\vu$ with precision $\tau_\z=\gamma_s=\alpha_s^2/\sigma_s^2$, and Stable unCLIP is a conditional model of the form in \autoref{eq:direct-cond-train}. Since $\tilde\z$ depends on the prompt only through $\x$, \autoref{eq:sem-noise-posterior} holds with $q(\x\mid\mathrm{prompt})$ in place of $q(\x)$. Conditioning on $\vu_s=\alpha_s\vu^\star$, where $\vu^\star$ is the standardized embedding of a reference image, sets $\tilde\z=\gamma_s\vu^\star$, so that, with $\gamma=\gamma_s$, the model samples from the tilt by $r(\x)=-\tfrac{\gamma}{2}\|\vphi(\x)-\vu^\star\|^2$,
\begin{equation}\label{eq:unclip-tilt}
    q(\x\mid\mathrm{prompt},\tilde\z=\gamma\vu^\star,\tau_\z=\gamma)\propto q(\x\mid\mathrm{prompt})\exp\bigl(-\tfrac{\gamma}{2}\|\vphi(\x)-\vu^\star\|^2\bigr).
\end{equation}
We map each $\gamma$ to the level whose $\gamma_s$ is nearest on a logarithmic scale. The last level gives $\gamma\approx2\cdot10^{-9}$, which we denote by $\gamma=0$, and the first gives $\gamma\approx2.4\cdot10^{4}$, close to the default use of unCLIP with the clean reference embedding.

\paragraph{Sampling and guidance.} We sample with 25 steps of DPM-Solver++ \citep{lu2025dpmsolverpp}. The Stable unCLIP pipeline in diffusers applies classifier-free guidance with the image embedding set to zero in the unconditional branch, which multiplies the reward by the guidance weight $w$. We instead keep the reference condition in both branches, so that guidance targets
\begin{equation*}
    q(\x\mid\tilde\z,\tau_\z)^{1-w}\,q(\x\mid\mathrm{prompt},\tilde\z,\tau_\z)^{w}\propto q(\x)^{1-w}q(\x\mid\mathrm{prompt})^{w}\exp\bigl(r(\x)\bigr),
\end{equation*}
the tilt of the guided text-conditional model with the same strength $\gamma$ for every $w$.

\begin{figure}[h]
    \centering
    \includegraphics[width=\linewidth]{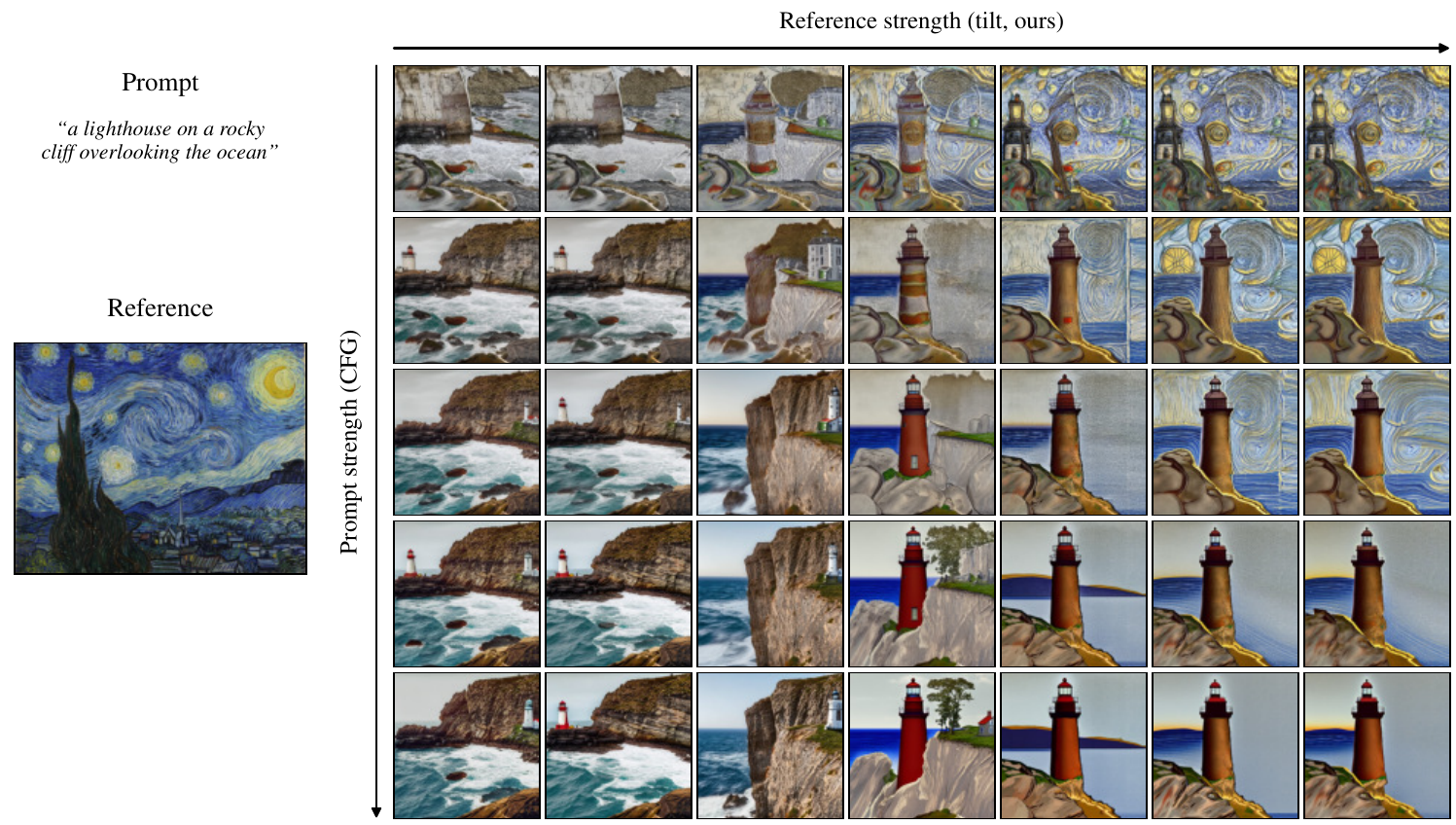}
    \caption{\textbf{Tilting Stable unCLIP toward a reference image without retraining.} Columns use $\gamma=0$, $0.1$, $0.3$, $1$, $3$, $10$ and $2.4\cdot10^{4}$, and rows use guidance weights $w=1$, $3$, $5$, $7.5$ and $10$. All images share the initial noise.}
    \label{fig:unclip-grid}
\end{figure}

\paragraph{Effect of $\gamma$ and $w$.} The images change mostly between $\gamma=0.1$ and $\gamma=10$ (\autoref{fig:unclip-grid}). Higher guidance weights reduce the influence of the reference, since guidance sharpens the prompt conditional but leaves the reward unchanged, and $w=1$ gives low-quality images. We use $w=5$ in the remaining experiments.

\begin{figure}[h]
    \centering
    \includegraphics[width=\linewidth]{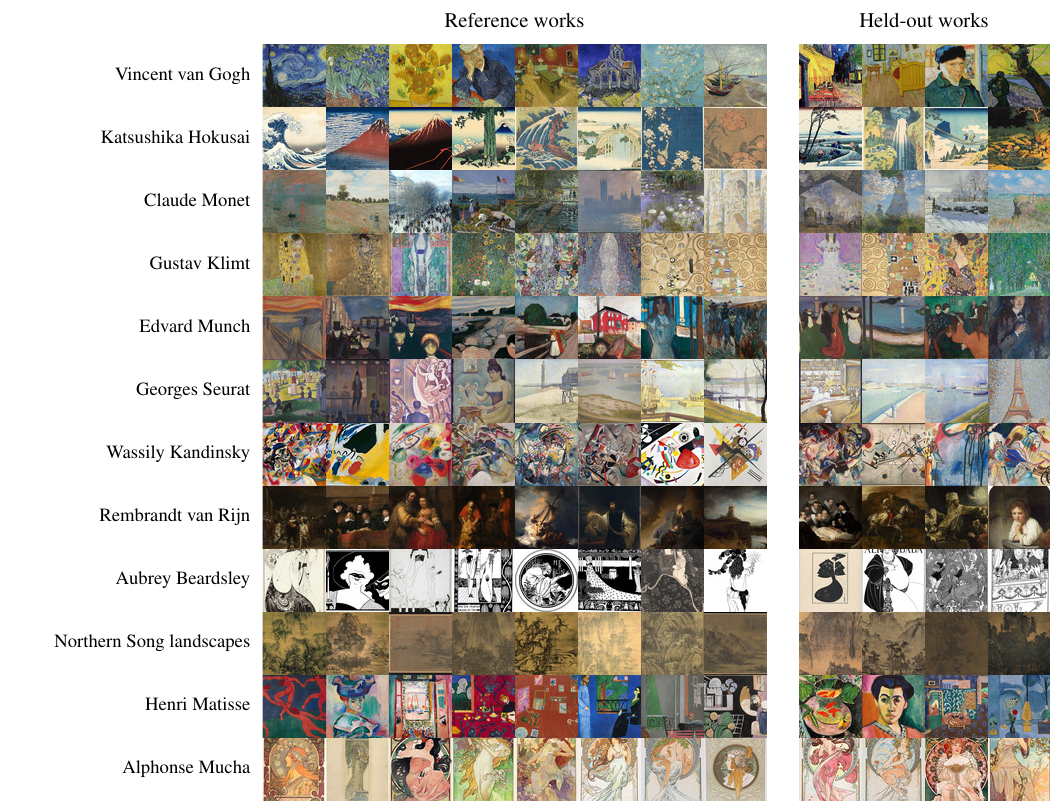}
    \caption{\textbf{Styles of the evaluation.} Each row shows the eight reference works of a style, starting with the painting used for single-reference tilting, and the four held-out works used only to measure style, as the square center crops seen by the encoders.}
    \label{fig:unclip-styles}
\end{figure}

\paragraph{Quantitative evaluation.} We pair 12 scene prompts written without describing a style (\autoref{tab:unclip-prompts}) with 12 public-domain paintings, the first reference work of each style in \autoref{fig:unclip-styles}, and generate three seeds for each of the 144 pairs. We measure how closely an image follows its prompt by the cosine similarity between the SigLIP 2 \citep{tschannen2025siglip} embeddings of the image and the prompt, which we call prompt similarity. We measure how strongly the image takes on the style of the painting by the cosine similarity between their CSD \citep{somepalli2024measuring} embeddings, which we call style similarity. CSD is an image encoder trained to compare artistic styles. Both SigLIP 2 and CSD differ from the CLIP encoder that defines the reward. We also measure diversity by the mean pairwise cosine distance between the DINOv2 embeddings of the images generated with the three seeds of a pair. The Kandinsky 2.2 experiments (\appref{app:kandinsky}) use the same prompts and paintings.

\begin{table}[h]
    \centering
    \small
    \caption{Prompts of the quantitative evaluation.}
    \label{tab:unclip-prompts}
    \begin{tabular}{ll}
        \toprule
        a lighthouse on a rocky cliff overlooking the ocean & a sailboat on a calm lake at sunset \\
        a cat sitting on a windowsill & a fox in a snowy forest \\
        a busy city street at night & a medieval castle on a hill \\
        a mountain village in winter & a steam train crossing a viaduct \\
        a bowl of fruit on a wooden table & a market square with people and stalls \\
        a portrait of an old fisherman & a dog running on a beach \\
        \bottomrule
    \end{tabular}
\end{table}

\begin{figure}[h]
    \centering
    \includegraphics[width=0.75\linewidth]{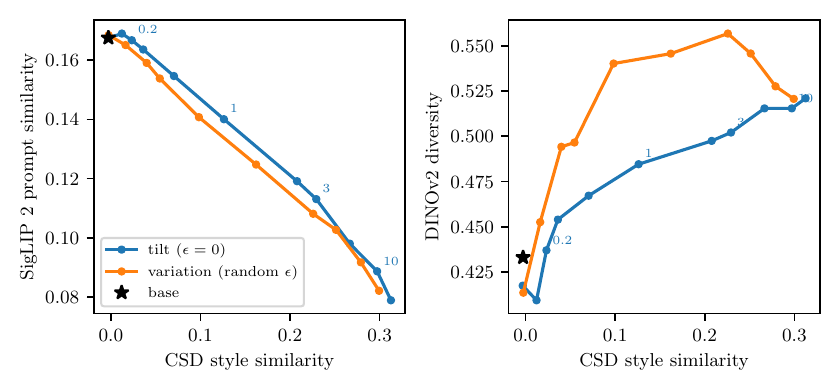}
    \caption{\textbf{Tilt and unCLIP image variations on 144 prompt--painting pairs at $w=5$.} Prompt similarity and diversity against style similarity to the reference, for the tilt ($\eps=0$) and the image variations with random $\eps$ at $\gamma$ from 0 to 10, with the tilt also at $\gamma=2.4\cdot10^{4}$. The star marks the model conditioned on the zero embedding that unCLIP uses for embedding dropout. Numbers give $\gamma$ for the tilt.}
    \label{fig:unclip-tradeoff}
\end{figure}

\paragraph{Tilt and unCLIP image variations.} Stable unCLIP generates image variations by conditioning on $\vu_s=\alpha_s\vu^\star+\sigma_s\eps$ with random $\eps$ \citep{rombach2023stableunclip}. This sets $\tilde\z=\gamma_s(\vu^\star+\eps/\sqrt{\gamma_s})$, which gives the tilt toward the random target $\vu^\star+\eps/\sqrt{\gamma_s}$ rather than $\vu^\star$. At the same style similarity, image variations are more diverse than the tilt and have a slightly lower prompt similarity (\autoref{fig:unclip-tradeoff}), so randomizing the target adds diversity at a small cost in prompt similarity.

\begin{figure}[h]
    \centering
    \includegraphics[width=\linewidth]{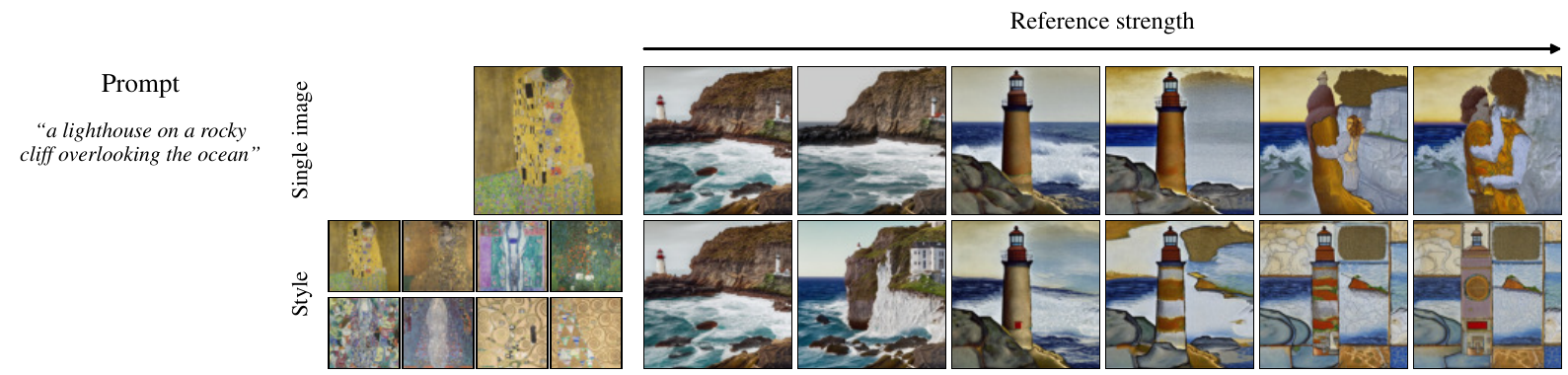}
    \caption{\textbf{Tilting toward a single painting and toward a style.} Top: tilt toward The Kiss. Bottom: tilt toward the rescaled mean of the embeddings of eight works by Klimt. The first column conditions on the zero embedding, and the remaining columns use $\gamma=0.3$, $1$, $3$, $10$ and $2.4\cdot10^{4}$ at $w=5$, with shared initial noise. At strong tilts, the single-painting tilt replaces the lighthouse with the couple from The Kiss, whereas the style tilt keeps the lighthouse.}
    \label{fig:unclip-set}
\end{figure}

\paragraph{Tilting toward a style.} The product of tilts toward references $\vu_1,\dots,\vu_N$, each with strength $\gamma/N$, is the tilt toward their mean $\bar\vu$ with strength $\gamma$, since
\begin{equation*}
    \sum_{i=1}^{N}\frac{\gamma}{2N}\|\vu-\vu_i\|^2=\frac{\gamma}{2}\|\vu-\bar\vu\|^2+\mathrm{const}.
\end{equation*}
Conditioning on $\bar\vu$ therefore tilts toward a style rather than a single painting. We test this on 12 styles, each with eight reference works (\autoref{fig:unclip-styles}), and measure style similarity to four held-out works of the same style, so that copying a reference does not count as matching the style. At the same style similarity, the tilt toward $\bar\vu$ has a higher prompt similarity than tilts toward single references, but it does not reach the style similarity of strong single-reference tilts. Since averaging reduces the norm, we also rescale $\bar\vu$ to the average norm of single embeddings, a heuristic that no longer corresponds to a product of tilts. The rescaled mean reaches a higher style similarity than single references while keeping the higher prompt similarity (\autoref{fig:unclip-set}).

\section{Tilting Kandinsky 2.2}\label{app:kandinsky}
Following the factorized construction of \autoref{sec:embedding-only-new}, we tilt the pretrained prior of Kandinsky 2.2 (\autoref{sec:kandinsky-exp}) and compare the tilt with interpolation between the sampled and the reference embeddings, a common way to combine a prompt with a reference image in models with a semantic prior \citep{ramesh2022hierarchical}.

\paragraph{Model and sampling.} The prior of the Kandinsky 2.2 model \citep{razzhigaev2023kandinsky} is a text-conditional diffusion model over the CLIP ViT-bigG/14 image embedding, standardized with fixed per-dimension statistics to $\z=\vphi(\x)$. A separate decoder generates an image from the unstandardized embedding and does not receive the prompt, so the model samples
\begin{equation*}
    p(\x\mid\mathrm{prompt})=\int p(\x\mid\z)\,q(\z\mid\mathrm{prompt})\,d\z.
\end{equation*}
For the reward $r(\z)=-\tfrac{\gamma}{2}\|\z-\z^\star\|^2$, we replace $q(\z\mid\mathrm{prompt})$ by $q^r(\z\mid\mathrm{prompt})\propto q(\z\mid\mathrm{prompt})\exp r(\z)$ and keep $p(\x\mid\z)$ unchanged, as in \autoref{sec:embedding-only-new}. By \autoref{eq:general-denoiser-new}, the tilted prior denoiser is $D(\tilde\z+\gamma\z^\star,\tau+\gamma)$, where $(\tilde\z,\tau)$ are the natural coordinates of the current state (\autoref{sec:diffusion-natcoord-new}). At each step, we map the shifted coordinates to the state and noise level of the model, interpolating the level between entries of the discrete schedule linearly in $\log\tau$, evaluate the prior once, and take the scheduler step from the current state. We use 25 prior steps without classifier-free guidance, instead of the default weight of 4, so that the tilted denoiser targets $q^r(\z\mid\mathrm{prompt})$. The decoder uses 30 steps, a guidance weight of 4 and a resolution of $512\times512$.

\paragraph{Interpolation baseline.} Interpolation starts from the untilted prior sample $\z_{\mathrm{base}}$ with the same prompt and noise. Linear interpolation uses $(1-\rho)\z_{\mathrm{base}}+\rho\z^\star$. Spherical interpolation acts on the unstandardized embeddings, interpolating their directions along a great circle and their norms linearly. At $\rho=1$, both decode the reference embedding. All conditions of a prompt and seed share the prior and decoder noise.

\paragraph{Evaluation.} We use the 144 prompt--painting pairs and three seeds of \appref{app:unclip}, with 14 values of $\gamma$ between 0.08 and 1 and 10 values of $\rho$ between 0.3 and 1. Since $\gamma$ and $\rho$ are not comparable, we compare the methods at matched values of a mean statistic, interpolating each curve linearly between strengths.

\paragraph{Embedding space.} Among all distributions with a given expected reward, the tilt has the smallest KL divergence from $q(\z\mid\mathrm{prompt})$ (\autoref{eq:tilt-objective}). At the same mean distance to $\z^\star$, tilted embeddings should therefore look more like samples from the prior than interpolated ones. As a simple check, we compare the norms of the sampled standardized embeddings with the range of norms of untilted samples, at three matched mean distances to $\z^\star$. At most 7\% of the tilted embeddings fall outside this range, compared with all linearly interpolated embeddings, whose norm shrinks, and about 80\% of spherically interpolated ones, which preserve the norm of the unstandardized embedding but not of the standardized one. Staying within this range is necessary but not sufficient for being typical of the prior.

\begin{figure}[h]
    \centering
    \includegraphics[width=0.75\linewidth]{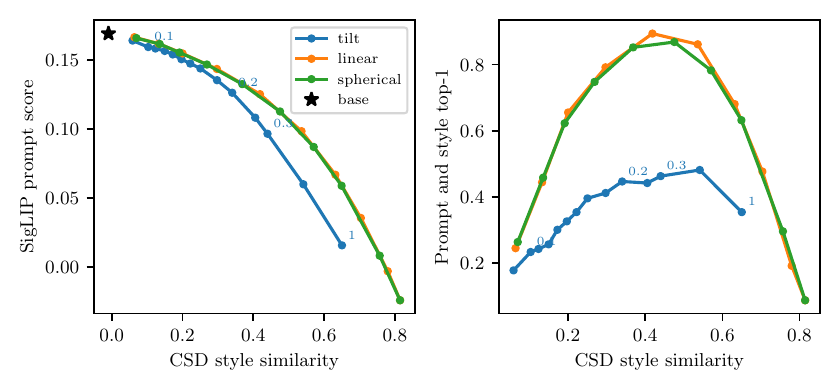}
    \caption{\textbf{Tilt and interpolation on Kandinsky 2.2 for 144 prompt--painting pairs with three seeds each.} SigLIP 2 prompt similarity and the fraction of images whose prompt and painting are both the most similar of the 12 under SigLIP 2 and CSD, against CSD style similarity to the reference. The star marks the untilted prior. Numbers give $\gamma$ for the tilt.}
    \label{fig:kandinsky-tradeoff}
\end{figure}

\paragraph{Decoded images.} In this experiment, staying closer to the prior does not improve the decoded images, and interpolation gives a better trade-off between style and prompt similarity than the tilt (\autoref{fig:kandinsky-tradeoff}). We measure prompt similarity with SigLIP 2 and style similarity with CSD as in \appref{app:unclip}. At the same mean CSD, the prompt similarity of the tilt does not differ significantly from that of linear or spherical interpolation up to a CSD of 0.3, and is lower by 0.04 at a CSD of 0.6 (95\% bootstrap interval over prompts and paintings $[0.02,0.06]$). To check whether an image shows the prompt and the style at the same time, we also count how often its own prompt is the most similar of the 12 prompts under SigLIP 2 and its own painting the most similar of the 12 paintings under CSD. At its best strength, this holds for 48\% of the tilted images, compared with 89\% for linear and 87\% for spherical interpolation, since a tilted image tends to either keep the prompt or reproduce the painting.

\begin{figure}[h]
    \centering
    \includegraphics[width=\linewidth]{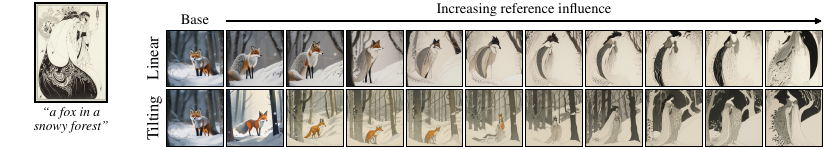}\\[2mm]
    \includegraphics[width=\linewidth]{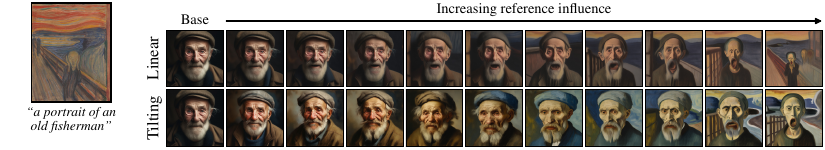}\\[2mm]
    \includegraphics[width=\linewidth]{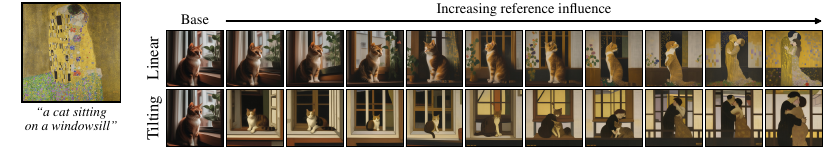}
    \caption{\textbf{Tilt and linear interpolation on Kandinsky 2.2.} The first column shows the untilted sample. Linear interpolation uses $\rho$ from 0.2 to 1. The tilt uses ten values of $\gamma$ up to 1, chosen separately for each example to show the transition, so the columns of the two rows are not matched in style similarity. All images of an example share the prior and decoder noise.}
    \label{fig:kandinsky-pairs}
\end{figure}

\paragraph{Dependence on $\gamma$.} With the prior and decoder noise fixed, the samples do not change smoothly with $\gamma$. In the examples of \autoref{fig:kandinsky-pairs}, the sampled embedding barely changes up to some $\gamma$ and then moves by roughly half of its distance to $\z^\star$ within a narrow range of $\gamma$, for the fox between $\gamma=0.045$ and $0.05$. The noise level at which we query the prior changes continuously with $\gamma$, so these jumps do not come from the discrete schedule of the prior. The mean CSD still increases smoothly with $\gamma$ (\autoref{fig:kandinsky-tradeoff}) because the jumps occur at different $\gamma$ for different pairs and seeds. Interpolation instead starts from the untilted sample, and its images change gradually with $\rho$.

\end{document}